\documentclass[10pt,twocolumn,letterpaper]{article}
\usepackage[T1]{fontenc}

\usepackage[pagenumbers]{cvpr} 

\usepackage{comment}
\usepackage[dvipsnames]{xcolor}

\usepackage{amsthm}
\newtheorem{theorem}{Theorem}

\usepackage{adjustbox}
\usepackage{colortbl}
\usepackage{multirow}
\usepackage{svg}
\usepackage{graphicx}
\usepackage{diagbox}
\usepackage{url}

\definecolor{cvprblue}{rgb}{0.21,0.49,0.74}
\usepackage[pagebackref,breaklinks,colorlinks,citecolor=cvprblue]{hyperref}

\usepackage{parskip}
\newcommand{\setParDis}{\setlength {\parskip} {0.1pt} }

\def\confName{CVPR}
\def\confYear{2024}

\title{Improved Implicit Neural Representation \\ with Fourier Reparameterized Training}

\author{Kexuan Shi \quad Xingyu Zhou \quad Shuhang Gu\thanks{Corresponding author}\\
School of Computer Science and Engineering, UESTC\\
{\tt\small \{kexuanshi712, xy.chous526, shuhanggu\}@gmail.com}
}

\begin{document}
 \setParDis
 \maketitle
\begin{abstract}

\hspace{1pc}Implicit Neural Representation (INR) as a mighty representation paradigm has achieved success in various computer vision tasks recently. 
Due to the low-frequency bias issue of vanilla multi-layer perceptron (MLP), existing methods have investigated advanced techniques, such as positional encoding and periodic activation function, to improve the accuracy of INR.
In this paper, we connect the network training bias with the reparameterization technique and theoretically prove that weight reparameterization could provide us a chance to alleviate the spectral bias of MLP.
Based on our theoretical analysis, we propose a Fourier reparameterization method which 
learns coefficient matrix of fixed Fourier bases to compose the weights of MLP.
We evaluate the proposed Fourier reparameterization method on different INR tasks with various MLP architectures, including vanilla MLP, MLP with positional encoding and MLP with advanced activation function, etc.
The superiority approximation results on different MLP architectures clearly validate the advantage of our proposed method.
Armed with our Fourier reparameterization method, better INR with more textures and less artifacts can be learned from the training data. The codes are available at \url{https://github.com/LabShuHangGU/FR-INR}.


\begin{figure}
    \includegraphics[width=\linewidth]{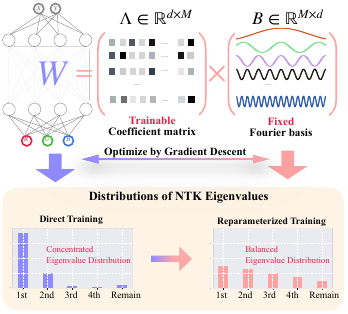}
    \vspace{-5mm}
    \caption{A conceptual illustration of our Fourier reparameterization method. We reparameterize the linear weight $\mathbf{W}$ with a trainable coefficient matrix $\mathbf{\Lambda}$ and a fixed Fourier basis matrix $\mathbf{B}$.
    More balanced eigenvalue distribution of neural tangent kernel (NTK) matrix implies that our method is able to alleviate the low-frequency bias of deep neural network, and therefore leads to better implicit neural representation.}
    \label{fig:fourier_structure}
    \vspace{-3mm}
\end{figure}
\end{abstract}    
\section{Introduction}
\hspace{1pc}Recently, a novel signal representation paradigm called implicit neural representation (INR) has gained great attention in the field of computer vision and graphics. 
The main idea of INR is 
using multi-layer perceptron (MLP) to
parameterize continuous and differentiable functions in an implicit manner.
For example, given a gray scale image, INR takes the coordinates of the pixels as inputs to MLP and trains it to output the exact gray values using gradient-based optimization methods.
Benefiting from the continuity nature and the expressive power of MLP, a more versatile continuous representation can be learned than traditional discrete grid-based methods.
Consequently, INR has achieved state-of-the-art performance across a variety of tasks such as signal representation \cite{saragadam2023wire,sitzmann2020implicit}, 3D shape reconstruction \cite{3D_1,3D_2,3D_3} and novel view synthesis \cite{nerf1,nerf2,nerf3}.

\hspace{1pc}Despite the universal approximation capabilities of MLP which have been proved in \cite{universal}, obtaining highly accurate INR is not trivial.
Specifically,
MLP with ReLU activation function often fails to represent
the high-frequency components of the signal such as the complex texture information in images and the intricate geometric shapes involved in 3D shape reconstruction.
Such tendency of MLP to learn simple patterns of the target function is referred to as spectral bias \cite{Rahaman} or low-frequency bias \cite{xu_fourier_framework}. 
To improve the performance of INR, great efforts have been made to alleviate or circumvent the spectral bias of MLP. 
One main category of approaches find that the difficulty of learning high frequencies becomes easier by increasing the complexity of the input data manifold \cite{Rahaman} and explicitly extracting high-frequency features (for example, positional encoding \cite{nerf1}) to deal with the spectral bias issue.
While, inspired by the pioneer work of \cite{sitzmann2020implicit}, another main class of approaches exploit advanced activation functions \cite{yuce2022structured, saragadam2023wire,fathony2020multiplicative} for pursuing more accurate approximation.
%

\hspace{1pc}In this paper, 
we dive into details of the low-frequency bias issue and prove that reparameterizing weights of MLP with appropriate bases could provide us with a chance to narrow the gap between the magnitude of low-frequency  and high-frequency loss, \textit{i.e.} to alleviate the low-frequency bias during the training of deep neural networks.
We propose a Fourier reparameterization strategy (see Fig. \ref{fig:fourier_structure}) and evaluate our method on simple 1D function approximation task and several real-world vision applications.
Experimental results clearly validate our advantages in alleviating the low-frequency bias issue for improving approximation accuracy.
Our study provides the literature with a practical reparameterization solution for improving the approximation accuracy of MLP without modifying its network architecture, previously, which so far has only been studied for the convolutional architectures.
Moreover, our theoretical analysis sheds new light on the advantage of reparameterized training by connecting it with the low-frequency bias issue.
We hope our study could inspire future works in improving training dynamics of deep neural networks with sophisticated reparameterization methods.

\hspace{1pc}Our contributions are summarized as follows:
\begin{itemize}
    \item We connect network training bias  with  reparameterization technique and theoretically prove that 
    appropriate reparameterization could alleviate the low-frequency bias issue by altering the magnitude of gradients from different frequencies.
    \item We propose a practical reparameterization method for multi-layer perceptron, \textit{i.e.} the Fourier reparameterization scheme, which could effectively improve the approximation accuracy of implicit neural representation without modifying its network architecture.
    \item We provide detailed experimental analysis on a wide range of implicit neural representation tasks. Our Fourier reparameterization method allows the improvement of commonly used network architectures and provides an implicit neural representation with more high-frequency details.
\end{itemize}

 \section{Related Work}
\label{sec:methodology}

\textbf{Implicit neural representations.}
The novel signal representation paradigm representing a signal as an implicit continuous function by neural networks has gained lots of attention. 
Recent works have demonstrated the remarkable performance and memory-efficient property in many representation tasks such as 2D image representation \cite{saragadam2023wire, sitzmann2020implicit, Imagefitting_1,fathony2020multiplicative,guass}, occupancy volume representation \cite{3D_1,3D_5,3D_6,3D_8}, view synthesis \cite{nerf1,nerf2,nerf3,nerf4,nerf5,nerf6} and virtual reality \cite{VR}. 
 However, the popular activation function ReLU empirically can't achieve the satisfactory performance with vanilla MLP. 
 Therefore, various modifications have been studied. 
 The mainstream modifications can be classified into two aspects. 
 The first aspect focuses on the input domain. 
 Rahaman \etal \cite{Rahaman} show that learning high frequency components gets easier with the increasing of the complexity of the input data manifold. Inspired by this phenomenon, Mildenhall \etal \cite{nerf1} use the positional encoding which adopts the sinusoidal mapping of the input features as the new input in the view synthesis and achieve remarkable performance. Zhong \etal \cite{PE2} also use this method to reconstruct more accurate continuous distributions of 3D protein structure. Recent studies \cite{param-encoding2,param-encoding3,param-encoding4} propose to encode input coordinates by learned features. Inspired by this, Xie \etal \cite{diner} rearrange the order of input coordinates and obtain more low-frequency components, avoiding confrontation with spectral bias.
 The second aspect focuses on the activation function. Sitzmann \etal \cite{sitzmann2020implicit} find that using periodic functions, such as the sinusoidal function, as activation functions can achieve remarkable performance in signal fitting. Sinusoidal functions are also explored by Fathony \etal \cite{fathony2020multiplicative} in multiplicative filter networks. Y\"{u}ce \etal \cite{yuce2022structured} explain the success of the usage of  periodic function from a structured dictionary perspective.  
 Inspired by this perspective, Saragadam \etal \cite{sitzmann2020implicit} use a complex Gabor wavelet activation 
 and achieve robust and accurate representations.

\vspace{\baselineskip}
\noindent{\textbf{Spectral bias.}}
The term spectral bias \cite{Rahaman} also known as the low-frequency bias \cite{xu_fourier_framework} implies that MLP tends to learn simple patterns of the real data \cite{Arpit} or the low-frequency components of the target function \cite{xu_fourier_framework}.
Arpit \etal \cite{Arpit} first find this phenomenon which has attracted lots of follow up studies.
%
%
Rahaman \etal \cite{Rahaman} exploit the structure of ReLU networks to evaluate its Fourier spectrum and estimate the relationships between the spectral norm of MLP weights and the amplitude of the output of MLP at different frequencies. 
Xu \cite{xu_fourier_framework} builds a theoretical framework by Fourier Analysis to decompose gradients in the frequency domain and discusses the distributions of the absolute gradients of the parameters at different frequencies. 
From the perspective of the neural tangent kernel (NTK) theory \cite{NTK1}, components of the target function corresponding to larger kernel eigenvalues will be learned faster \cite{NTK3,NTK4}. Therefore, Tancik \etal \cite{tancik2020fourier} propose to leverage the eigenvalues of the NTK matrix to analyze the spectral bias of MLP. In this paper, we dive into details of the low-frequency bias issue and find that network reparameterization could provide us a chance to alleviate the bias in network training.
We analyze our proposed method with the frequency decomposed loss, gradient \cite{xu_fourier_framework} and NTK theory \cite{tancik2020fourier}.
Our experimental results validate our idea of improving low-frequency bias with weight reparameterization.

\vspace{\baselineskip}
\noindent{\textbf{Weight reparameterization.}}
In order to reduce the inference cost of deep learning models \cite{multi-branch_1,multi-branch_2}, weight reparameterization method is proposed by Zagoruyko \etal \cite{diracnets}.
Inspired by this work, various reparameterization methods have been explored to train networks with mergeable auxiliary structures \cite{ding2019acnet, ding2021repvgg, park2017design, xie2022fourier,reparameter,ding2021resrep}.
Recently, there are still researches which exploit the idea of reparameterization to design network optimizer for advanced training \cite{ding2022re}. 
In this paper, we adopt the idea of weight reparameterization to improve the approximation accuracy of MLP.
For the first time, we link the reparameterization technique with network training bias and theoretically prove the possibility of mitigating frequency-bias with reparameterized training.
We hope our theoretical analysis could shed new light on network reparameterization and inspire future studies on advanced reparameterization method.

 \section{Methodology}
\subsection{The formulation of INRs}
\hspace{1pc}The task of INR is to approximate a target function   with a multi-layer perceptron (MLP): $f_\mathbf{\Theta}(x)\approx g(x)$;
where $g(x): \mathbb{R}^{d_0} \mapsto \mathbb{R}^{d_{N}}$ is the target function which defines a mapping from a $d_0$-dimensional real space to a $d_{N}$-dimensional real space, and $f_{\mathbf{\Theta}}(x)$ is  a N-layer MLP with the learnable parameters set $\mathbf{\Theta}$.
Denote the output of $n$-th layer as:
\begin{equation}
\mathbf{y^{(n)}} = \sigma(\mathbf{W^{(n)}y^{(n-1)}}+\mathbf{b^{(n)}}),
\end{equation}
where $\sigma$ is an element-wise nonlinear activation function; $\mathbf{W^{(n)}}\in \mathbb{R}^{d_{n}\times d_{n-1}}$ and $\mathbf{b^{(n)}}\in \mathbb{R}^{d_{n}}$ are the weight and bias for the $n$-th layer; $\mathbf{y^{(n)}}\in\mathbb{R}^{d_n}$ and $\mathbf{y^{(n-1)}}\in\mathbb{R}^{d_{n-1}}$ are the output and input of  the $n$-th layer, respectively. For $n=N$, we have that $\mathbf{y^{(N)}}=\mathbf{W^{(N)}y^{(N-1)}}+\mathbf{b^{(N)}}$.
The MLP parameters set $\mathbf{\Theta}= \{\mathbf{W^{(i)}},\mathbf{b^{(i)}}\}_{\mathbf{i=1,\dots,N}}$ is learned by minimizing the loss with gradient-based methods.

\hspace{1pc}The above idea of INR is memory efficient, and the continuous nature of MLP allows INR to model fine detail that is not limited by the grid resolution.
However, during the practical optimization process, gradients of parameters are dominated by the error of low-frequency components. Such a spectral bias hinders the accurate learning pace of MLP.  
Various modifications, including input feature adjustments \cite{tancik2020fourier,diner} and activation function adjustments \cite{sitzmann2020implicit,saragadam2023wire,guass,fathony2020multiplicative}, have been exploited for alleviating the low-frequency bias issue.
In this paper, we show that appropriate reparameterization of MLP is also beneficial for narrowing the gap between the gradient magnitude of high-frequency components and low-frequency components.

\subsection{Fourier reparameterization}
\label{fourier reparameterization}
\hspace{1pc}As we have introduced in the previous subsection, the weight matrix in the $n$-th layer of MLP is denoted as $\mathbf{W^{(n)}}\in \mathbb{R}^{d_{n}\times d_{n-1}}$.
Instead of directly calculating the gradient of $\mathbf{W^{(n)}}$ respect to the loss function, we reparameterize each row of $\mathbf{W^{(n)}}$ as a weighted combination of fixed Fourier bases:
\begin{equation}
    \mathbf{W^{(n)}}=\mathbf{\Lambda^{(n)}} \mathbf{B^{(n)}},
    \label{eq:rep}
\end{equation}
where $\mathbf{\Lambda^{(n)}} \in \mathbb{R}^{d_n\times M }$ are the coefficient matrix, 
and $\mathbf{B^{(n)}} \in \mathbb{R}^{M \times d_{n-1}}$ are $M$ Fourier bases.
Each Fourier basis is achieved by changing the frequency $\omega$ and phase $\varphi$ of a cosine function $cos(\omega z+\varphi)$:
\begin{equation}
\label{eq:basis}
    b_{i,j}= cos(\omega_{i} z_j+\varphi_i),for\ i=1,\dots,M;j=1,\dots,d_{n-1},
\end{equation}
where $\mathbf{z}=\{ z_j\}_{j=1,\dots,d_{n-1}}$ is the sampling position sequence. More implementation details will be introduced in section \ref{abla:length}.
Generally, we have $M\geq d_{n-1}$ which means that we reparameterize the weight matrix with over complete bases.

\hspace{1pc}Please note that in Eq. \ref{eq:rep}, each basis is with the same dimension as the input feature $\mathbf{y^{(n-1)}}$, which means that our reparameterization scheme firstly projects the input features onto a series of Fourier bases and weighted combines the projection coefficients to generate the input feature for the next layer:
 \begin{equation}
\mathbf{y^{(n)}} = \sigma(\mathbf{\Lambda^{(n)} B^{(n)} y^{(n-1)}}+\mathbf{b^{(n)}}).
\end{equation}
In the above equation, $\mathbf{B^{(n)}}$ is fixed and we only learn $\mathbf{\Lambda^{(n)}}$ during the training phase.
After training, we combine $\mathbf{\Lambda^{(n)}}$ and $\mathbf{B^{(n)}}$ to form the weight matrix $\mathbf{W^{(n)}}$ in the inference.
Therefore, our reparameterization approach only adjusts the training dynamic and will not affect the inference process of INR.
 Moreover, the proposed Fourier reparameterization approach does not affect the input feature space as well as nonlinear activation functions. Our method is compatible with existing techniques, including but not limited to positional encoding \cite{tancik2020fourier} and periodic activation functions \cite{sitzmann2020implicit}.

\subsection{Discussion}

\hspace{1pc}In this subsection, we analyze our reparameterization scheme in the frequency domain.
By carefully analyzing the gradients of learning parameters respect to different frequencies, we show that appropriate weight reparameterization provides us with a chance to alleviate the low-frequency bias in the network training.

\hspace{1pc}We start our analysis with the definition of some basic concepts \cite{xu_fourier_framework}.
We denote the Fourier Transform of the target function and the MLP representation of INR at frequency $k$ as: $\mathcal{F}[g](k)$ and $\mathcal{F}[f_{\mathbf{\Theta}}](k)$, respectively.
Then, the approximation error at frequency $k$ can be naturally achieved by $ E(k)=\mathcal{F}[g](k)-\mathcal{F}[f_{\mathbf{\Theta}}](k)$.
With $E(k)$, we further define the following notations: $E(k)=A(k)e^{i\theta (k)}$ and $\mathbb{L}(k)=|E(k)|^2$;
where $A(k)$ and $\theta(k) \in [\pi,\pi]$ indicate the amplitude and phase of $E(k)$; $\mathbb{L}(k)$ is the loss component of frequency $k$; $|\ \cdot \ |$ is the norm of the complex number.  

\hspace{1pc}Based on the above definitions, Xu \cite{xu_fourier_framework} shows that the spectral bias can be reflected on the absolute gradient values of parameters at different frequencies with the following Theorem 1:
\vspace{\baselineskip}
\begin{theorem}{(Theorem 1 in \cite{xu_fourier_framework})}
Consider a MLP with one hidden layer using tanh function $\sigma (x) $ as the activation function. For any frequencies $k_{1}$ and $k_{2}$ such that $k_{1} > k_{2} > 0$ and there exist $c_{1},c_{2},$ such that $A(k_{1})>c_{1}>0,A(k_{1})<c_{2}<\infty$,  we have
\begin{equation}
    \lim_{\delta \to 0}\frac{\mu (\{w_j:|\frac{\partial \mathbb{L}(k_2)}{\partial \Theta_{jl}}| > |\frac{\partial \mathbb{L}(k_1)}{\partial \Theta_{jl}}| for\ all\ j,l\} \cap B_{\delta})}{\mu (B_{\delta})} =1,
\end{equation}
where $B_{\delta }$ is a ball with radius $\delta$ centered at the origin and $\mu(\cdot)$ is the Lebesgue measure of a set and $\Theta_{jl}$ is a leanrable parameter in the parameters set.
\label{theorem1}
\end{theorem}
\hspace{1pc}Generally, Theorem 1 shows that in  the case of one hidden layer MLP, the gradient respect to low-frequency loss $\mathbb{L}(k_2)$ is almost always larger than the one respect to high-frequency part.
Although the condition of one hidden layer MLP is not appliable in most of practical cases, based on recent observations of low-frequency bias \cite{xu_fourier_framework,Rahaman,NTK1,Arpit}, the above Theorem 1 inspires us to assume such relationship to MLP with more hidden layers.
Then, we could have the following Theorem 2: 
\vspace{\baselineskip}
\begin{theorem}
Given a MLP with multiple hidden layers, reparameterize the weight matrix $\mathbf{W}\in \mathbb{R}^{d \times d}$ of one hidden layer with a trainable coefficient
matrix $\mathbf{\Lambda}\in \mathbb{R}^{d \times M}$ and the fixed basis matrix $\mathbf{B}\in \mathbb{R}^{M \times d} $. For any frequencies $k_1$ and $k_2$ such that $k_1>k_2>0$, given any $\epsilon \geq 0$ and fixed $i$, for $ j=1,2,\dots,M$, there must exist a set of basis matrices such that 

\begin{equation}
\fontsize{6.2}{0}\selectfont{|\frac{\partial \mathbb{L}(k_1)}{\partial \lambda_{ij}}/ \frac{\partial \mathbb{L}(k_2)}{\partial \lambda_{ij}}| \geq \max \{|\frac{\partial \mathbb{L}(k_1)}{\partial w_{i1}}/\frac{\partial \mathbb{L}(k_2)}{\partial w_{i1}}|,\dots,|\frac{\partial \mathbb{L}(k_1)}{\partial w_{id}}/\frac{\partial \mathbb{L}(k_2)}{\partial w_{id}}|\}-\epsilon},
\end{equation}
    \label{theorem2}
    where $\mathbf{W}(i,j)=w_{ij}$ and $\mathbf{\Lambda}(i,j)=\lambda_{ij}$.
\end{theorem}
\hspace{1pc}The detailed proof of Theorem 2 can be found in our supplementary file.
Theorem 2 implies that reparameterizing MLP weights with appropriate bases is able to enlarge the portion of high-frequency loss components in comparison to low-frequency  components, \textit{i.e.} improving the low-frequency bias in training MLP.
Although the optimal basis for frequency-bias adjustment is related to the training data and we are not able to achieve the optimal basis with negligible efforts,
we experimentally find that fixed Fourier basis 
is able to improve the low-frequency bias and provide better INR for various function approximation tasks.

\subsection{Implementation details}

\textbf{Basis construction.}
As we have introduced in section \ref{fourier reparameterization}, we establish Fourier bases with various frequency and phase parameters.
We adopt $P$ different phases and $2F$ different frequencies.
Concretely, the $\varphi$ in Eq. \ref{eq:basis} varies from $0$ to $2\pi(P-1)/P$ with step length $2\pi/P$;
for each phase value, we have a group of low-frequency bases with $\omega = \{1/F, 2/F, \dots, 1\}$ and a group of high-frequency bases with $\omega = \{1, 2, \dots, F\}$.
Based on the above basis construction scheme, we could obtain $M=2FP$ bases.
Details of the selected $F$ and $P$ values for different settings will be introduced in the experimental section.
We also provide ablation experiments in our supplementary field to analyze the effects of different design choices of $F$ and $P$.

\vspace{\baselineskip}
\noindent \textbf{Sampling strategy.} The cosine basis function used in Eq. \ref{eq:basis} is a continuous function.
We need to sample values from the continuous function to achieve discrete basis for weight reparameterization.
As the number of sampling points is determined by the neuron number of input feature, the only key hyper-parameter during the sampling process is the range of sampling.
To reflect characteristics of different frequency bases, we choose the maximum period ($T_{max} = 2\pi F $) of the adopted Fourier Bases as the sampling range.
To maintain the periodicity of bases, uniform sampling is employed.
Due to the symmetry of bases, we set the sampling interval as $[-\frac{1}{2}T_{max},\frac{1}{2}T_{max}]$.
The length of interval will be discussed in our ablation studies.

\vspace{\baselineskip}
\noindent \textbf{Initialization scheme.} Initialization techniques are one of the prerequisites for successfully training a deep neural network. 
The basic idea of the existing popular initialization strategies \cite{multi-branch_1,initial_2} is to let the network start in a regime with constant variance between inputs and outputs.
While our method reparameterizes the network weights as the fixed Fourier bases and trainable coefficients, initializing the learnable coefficients $\mathbf{\Lambda}$ with existing initialization techniques will deactivate the constant variance requirement.
We therefore adjust the initialization strategy for $\mathbf{\Lambda}$ to make the composed weight matrix have the same property as Kaiming initialized weights \cite{multi-branch_1}.
For ReLU activation function, we initialize the trainable coefficient matrix $\mathbf{\Lambda^{(n)}}$ using the following equation:
\begin{equation}
    \lambda^{(n)}_{ij} \sim U(-\sqrt{\frac{6}{M \sum_{t=1}^{d_{n-1}}{b^{(n)}_{jt}}^2}},\sqrt{\frac{6}{M \sum_{t=1}^{d_{n-1}}{b^{(n)}_{jt}}^2}}),
\end{equation}
where $\lambda_{ij}^{(n)}, b^{(n)}_{ij}$ is in the $i$-th row and $j$-th column of $\mathbf{\Lambda^{(n)}} \in \mathbb{R}^{d_n \times M}$  and $\mathbf{B^{(n)}}\in \mathbb{R}^{M \times d_{n-1}}$, respectively. For SIREN \cite{sitzmann2020implicit}, we have the similar initialization scheme. The detailed derivation can be found in our supplementary file.

\section{Experimental Analysis on Simple Function Approximation}
\hspace{1pc}We firstly conduct experiments on the simple function approximation task.
Thanks to the simplicity of the target function, we are able to analyze the behaviour of MLP with different techniques.
In the remaining of this section, we firstly introduce our experimental settings and then analyze the property of our proposed Fourier reparameterization (FR) in detail.

\subsection{Experimental settings}
\label{experimental settings}
\begin{figure}
    \centering
    \includegraphics[width=3.3in]{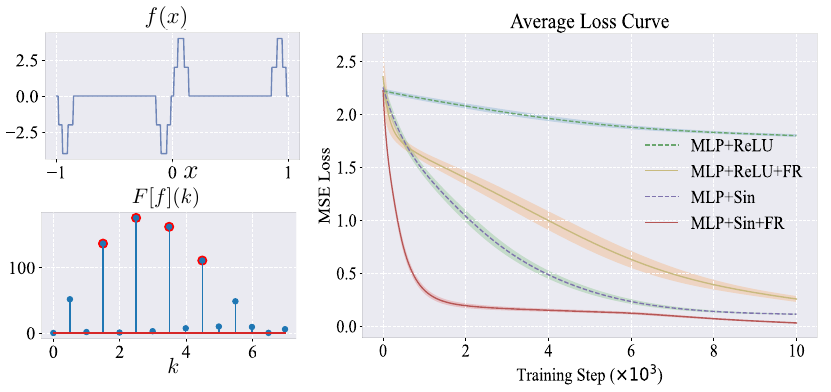}
    \caption{Visualization of simple function. The left side of the first row displays the visualization of the 1D simple function on the $x-y$ coordinate axis. The left side of the second row shows the amplitude of the function in the frequency domain. The right side presents the average loss curve with the shaded area indicating the fluctuation from 100 repetitions.}
    \label{fig:spectrum and loss}
\end{figure}

\hspace{1pc}In order to thoroughly  analyze the advantage of Fourier reparameterization, we establish a 1D function $f(x)$ by combining sine functions of different frequencies:
\begin{equation}
    2R(\frac{sin(3\pi x)\!+\!sin(5\pi x)\!+\!sin(7\pi x)\!+\!sin(9\pi x)}{2}),
\end{equation}
where $R( \cdot )$ is the rounding function for increasing the complexity of approximation. A similar rounded periodic function has been adopted in \cite{xu_fourier_framework} to analyze the spectral-bias of MLP.
A visualization of our adopted 1D function and its spectrum can be found in the first column of Fig. \ref{fig:spectrum and loss}. As designed by purpose, 
four distinct peaks marked by red dots can be observed in the stem plot of the spectrum.

\hspace{1pc}We utilize a four-hidden-layer MLP to approximate the $f(x)$ with 300 discrete values uniformly sampled in the interval $[-1,1]$, where each hidden layer consists of 128 neurons.
We conduct experiments on both the ReLU and the periodic activation function Sin.
The $\omega_0$ in Sin activation function is set as 5 for the pursuit of fast convergence \cite{sitzmann2020implicit}.
For our reparameterization approach, we reparameterize the weight matrices between consecutive hidden layers and set $F = 64$ and $P=16$. Therefore, we have $M = 2048$ bases in total.
The four comparison methods are denoted as: MLP+ReLU, MLP+ReLU+FR, MLP+Sin, MLP+Sin+FR. All the methods are trained with Adam optimizer \cite{Kingma2014AdamAM} for $10000$ iterations with a fixed learning rate 1e-6 and full-batch.

\subsection{Approximation results and analysis}
\hspace{1pc}The convergence curves by different methods can be found in Fig. \ref{fig:spectrum and loss}.
For both the ReLU and Sin activation function cases, 
Our FR approach improves the convergence speed as well as approximation error of vanilla MLP.
Especially for the naive MLP+ReLU case, training network with our proposed reparameterization method could improve the original training paradigm by a large margin.

\begin{figure*}
\begin{minipage}[c]{0.24\linewidth}
\centering
    \includegraphics[width=1.5in]{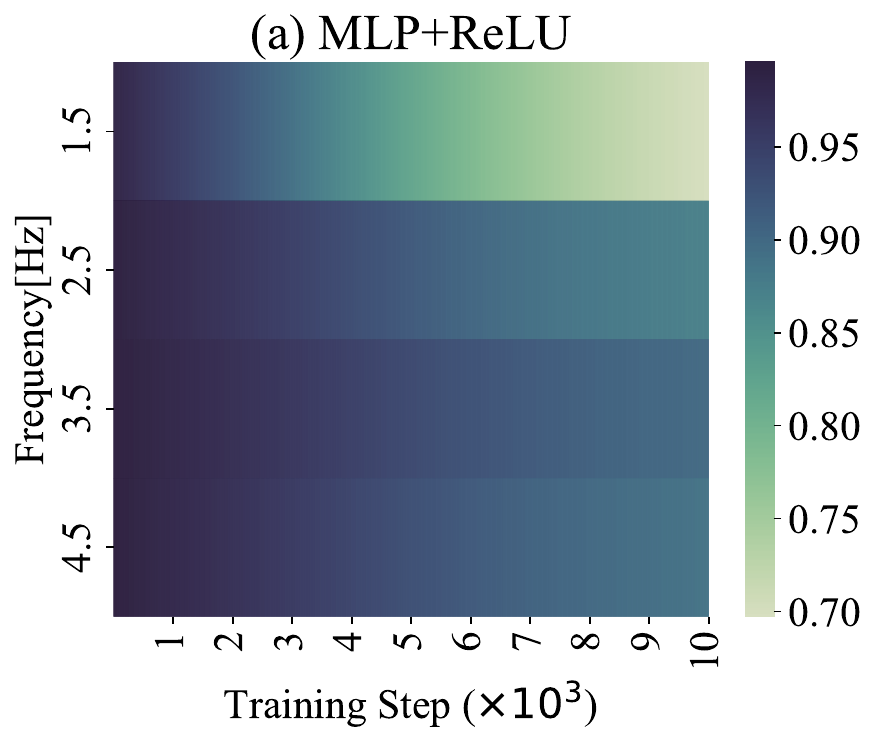}
\end{minipage}
\begin{minipage}[c]{0.24\linewidth}
\centering
    \includegraphics[width=1.5in]{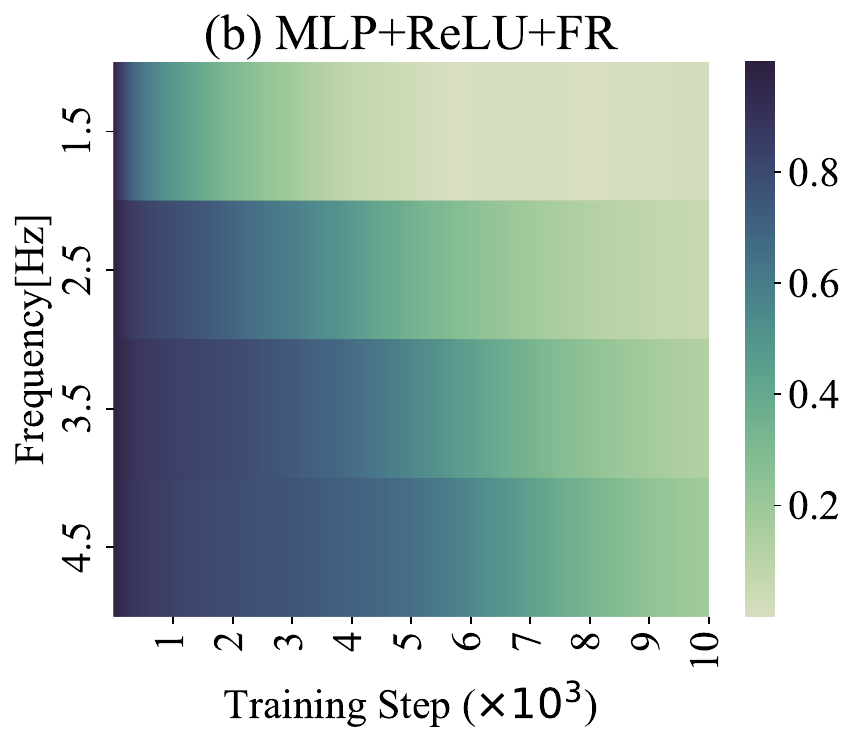}
\end{minipage}
\begin{minipage}[c]{0.24\linewidth}
\centering
    \includegraphics[width=1.5in]{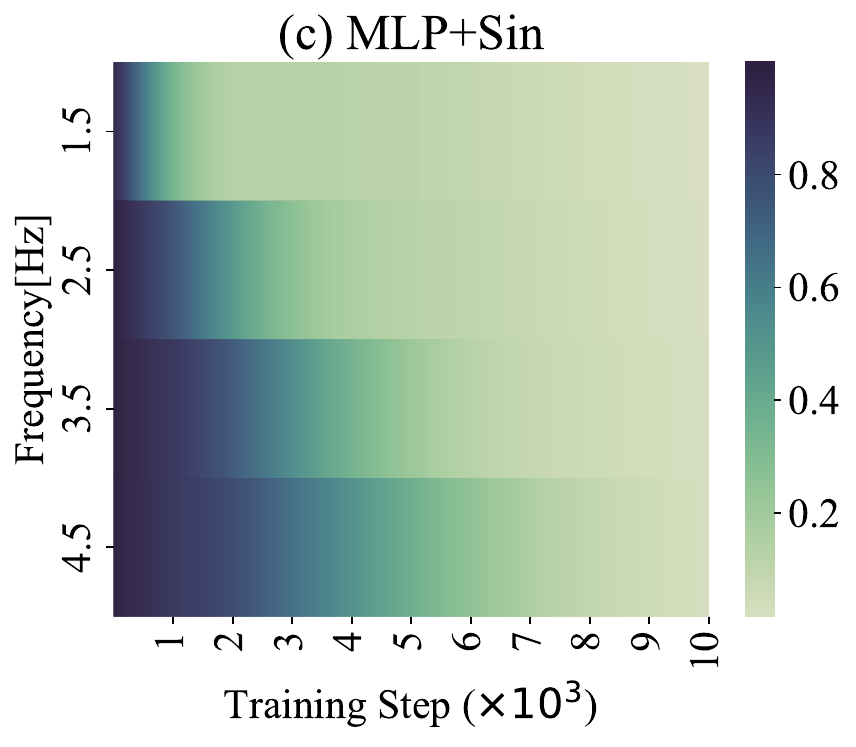}
\end{minipage}
\begin{minipage}[c]{0.24\linewidth}
\centering
    \includegraphics[width=1.5in]{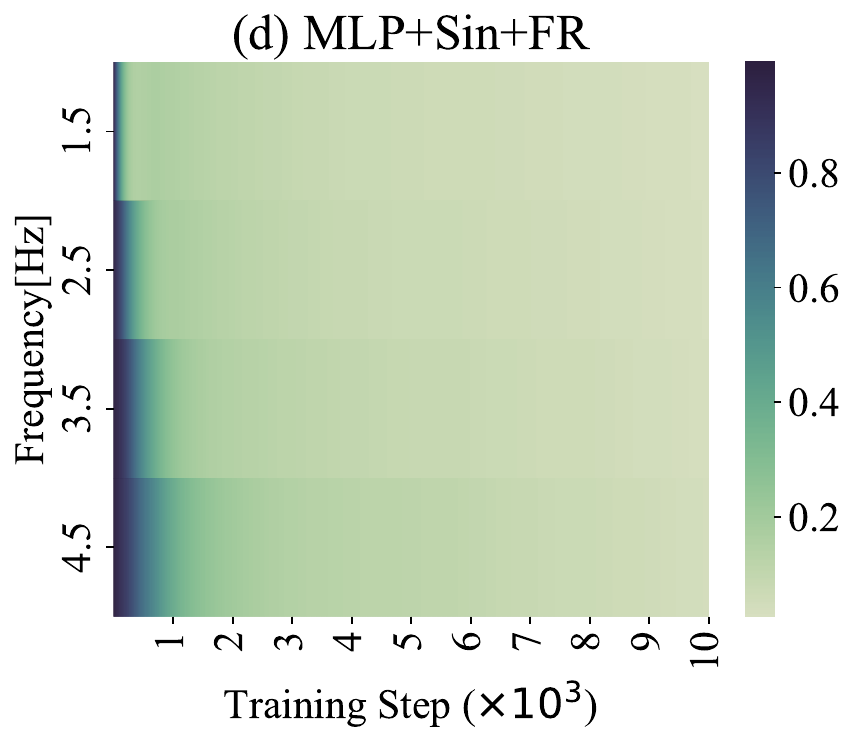}
\end{minipage}
\caption{Evolution of frequency-specific approximation error with training iterations of four different methods (x-axis for training step, y-axis for frequency and colormap for relative approximation error).}
\label{fig:FB}
\end{figure*}

\vspace{\baselineskip}
\noindent \textbf{Frequency-specific error analysis.}
In \cite{xu_fourier_framework}, using Discrete Fourier Transform, Xu \etal compute the relative difference $\Delta_k$ between the target signal and the output of MLP at frequency $k$ and empirically show the low-frequency bias of MLP:
\begin{equation}
\label{eq:FSEA}
    \Delta_k=\frac{|\mathcal{F}_D[g](k)-\mathcal{F}_D[f_\mathbf{\Theta}](k)|}{|\mathcal{F}_D[g](k)|},
\end{equation}
where $\mathcal{F}_D$ denotes the Discrete Fourier Transform.
We follow \cite{xu_fourier_framework} and use Eq. \ref{eq:FSEA} to analyze the frequency-specific approximation error after different numbers of iterations.
As can be found in Fig. \ref{fig:FB}, the evolution of frequency-specific error clearly shows the low-frequency bias in network training: 
the error of low-frequency components generally reduces much faster than that of high-frequency components.
The proposed FR method is able to narrow the gap between the dropping speed of low-frequency error and high-frequency error, thereby leading to overall faster convergence speed.

\vspace{\baselineskip}
\noindent \textbf{Neural tangent kernel.} 
Neural tangent kernel (NTK) \cite{NTK1} is a theory  to analyze the dynamic training process of neural networks.
Tancik \etal \cite{tancik2020fourier} have shown that components of the target function that correspond to larger kernel eigenvalues will be learned faster
and adopt kernel eigenvalues \cite{tancik2020fourier, NTK3, NTK4} to analyze the spectral bias. 
Since the conditions of the standard NTK theorey are not applicable to commonly used networks, we follow \cite{yuce2022structured, NTK_approximation} and utilize the following empirical NTK to analyze the training dynamics of different networks:
\begin{equation}
    k'_{NTK}(x_i,x_j)=J_{f_\mathbf{\Theta}}(x_i)J_{f_\mathbf{\Theta}}(x_j)^T,
\end{equation}
where $J_{f_\mathbf{\Theta}}(x_i)$ denotes the Jacobian matrix of the function $f_\mathbf{\Theta}$ at the $i$-th sample $x_i$ and $k'_{NTK}(x_i,x_j)$ is the element in the $i$-th row and $j$-th column of the empirical NTK matrix.

\begin{figure}
    \centering
    \includegraphics{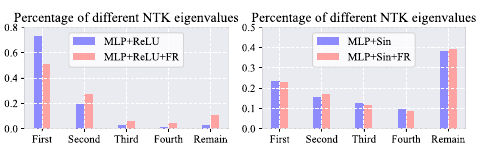}
        \caption{The magnitude of different NTK eigenvalues. 'First' denotes the percenatge of the  largest eigenvalue, 'Second' represents the percentage of the second-largest eigenvalue. 'Remain' refers to the percentage of the summation of remaining eigenvalues.}
       \label{fig:percentage of eigen}
\end{figure}

\hspace{1pc}The first four and the summation of the remaining eigenvalues by different methods are shown in Fig. \ref{fig:percentage of eigen}.
Consistent with the conclusion of \cite{tancik2020fourier}, the eigenvalues of MLP + ReLU decay rapidly, which means the model suffers severe spectral bias during training.
While, our Fourier reparameterization scheme is able to reduce the first eigenvalue and enlarge the other eigenvalues of the empirical NTK matrix, leading to more balanced eigenvalue distribution.

\section{Experimental Results on Vision Applications}
\hspace{1pc}Implicit neural representation has been utilized in different vision applications.
In this section, we evaluate the proposed Fourier reparameterization method on different vision applications.

\subsection{2D Color image approximation}
\label{2D_image}
\hspace{1pc}Natural images are extremely complex functions which simultaneously encompass rich low- and high-frequency components \cite{naturalimage}.
Single image fitting has become an ideal test bed \cite{sitzmann2020implicit,saragadam2023wire,diner} to evaluate the capability of implicit neural representation.
In our experiments, we attempt to parameterize
 a function $\phi: \mathbb{R}^2\mapsto \mathbb{R}^3$, $x \mapsto \phi(x)$ that represents a given discrete image in a continuous fashion.
\begin{figure}
\centering
    \includegraphics[width=\linewidth]{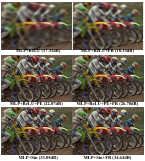}
    \caption{Visual examples of the 2D color image approximation results (PSNR) by different methods. Detailed experimental settings can be found in section \ref{2D_image}.}
    \label{Koda_show_fig}
\end{figure}

\begin{table*}
\scriptsize
\centering
\caption{Peak signal to noise ratio (PSNR) of 2D color image approximation results by different methods. Detailed experimental settings can
be found in section \ref{2D_image}. }
\begin{tabular}{l|cccccccc|c}
\toprule
Method & Kodim 01 & Kodim 02 & Kodim 03 & Kodim 04 & Kodim 05 & Kodim 06 & Kodim 07 & Kodim 08 & Average \\
\midrule
MLP + ReLU   & 19.37 & 26.12 & 25.11 & 24.57 & 17.31 & 21.69 & 20.79 & 15.68 & 21.33 \\
MLP + ReLU + FR & 20.34 & 26.58 & 27.21 & 25.72 & 18.33 & 22.25 & 22.47 & 16.64 & 22.44 \\
\midrule
MLP + ReLU + PE & 24.47 & 31.41 & 31.53 & 30.16 & 22.87 & 26.54 & 29.33 & 21.14 & 27.18 \\
MLP + ReLU + PE + FR & 27.64 & 33.92 & 34.45 & 33.23 & 26.78 & 29.83 & 34.13 & 24.70 & 30.59 \\
\midrule
MLP + Sin & 31.59 & 36.55 & 39.59 & 36.66 & 33.05 & 34.10 & 39.96 & 31.00 & 35.31 \\
MLP + Sin + FR & 33.45 & 38.68 & 39.58 & 37.96 & 34.64 & 34.45 & 39.76 & 32.16 & 36.34 \\
\bottomrule
\end{tabular}
\label{Table:Kodak24}
\end{table*}
\begin{figure*}
\centering
    \includegraphics[width=\linewidth]{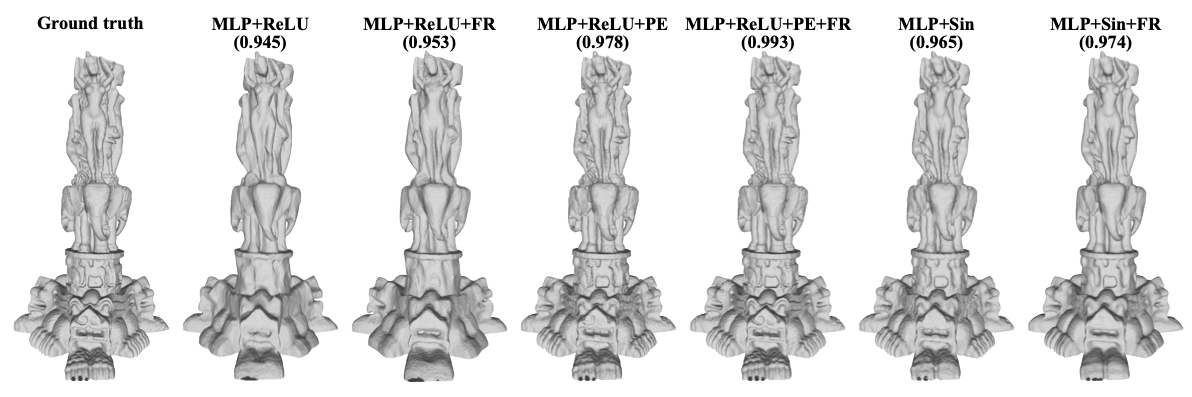}
    \caption{Visual examples of the shape representation results (IOU) by different methods. More experimental details can be found in section \ref{SDF}.}
    \label{Thai_state}
    \vspace{-3mm}
\end{figure*}

\hspace{1pc}We establish MLP with four hidden layers and each hidden layer contains 256 neurons.
We conduct experiments on three MLP architectures, i.e. (1)  MLP with Relu as the activation function (MLP+Relu), (2) MLP + Relu with Fourier positional encoding (MLP+Relu+PE) \cite{tancik2020fourier}, and (3) MLP with Periodic Sin activation function (MLP+Sin) \cite{sitzmann2020implicit}.
MLP+Relu+PE and MLP+Sin represent two important categories of techniques for improving INR. Experimental results on more activation functions and other input adjustments can be found in our supplementary file.
For each MLP architecture, we train baseline model which trains network parameters directly with the standard back-propagation approach and (+FR) model which utilizes our proposed Fourier reparameterization scheme in the training phase. We reparameterize the weight matrices between consecutive hidden layers and set $F=128,P=32$ for all the images in the experiment.
We use Adam optimizer to minimize the $\ell_2$ loss between ground truth pixel values and INR approximations.
The MLPs are trained with an initial learning rate of $10^{-4}$ for $3000$ iterations, and then we drop the learning rate to $10^{-5}$ and train the networks for another $7000$ iterations.
Full-batch training is adopted.

\hspace{1pc}In Table \ref{Table:Kodak24}, we report the PSNR values achieved by different INRs for approximating the first 8 images in the Kodak 24 dataset. 
Our FR method is able to improve  the approximation accuracy for all the three network architectures.
Some visual examples of the learned approximations can be found in Fig. \ref{Koda_show_fig}.
Our Fourier reparameterization  method enables the network to capture more fine details.

\subsection{Representing shapes with signed distance functions}

\label{SDF}
\hspace{1pc}Representing shapes with  differentiable signed distance functions (SDFs) has the advantage of 
modeling arbitrary topologies \cite{sitzmann2020implicit}. 
In this section, we evaluate the proposed Fourier reparameterization method on the shape representation task.
We follow the experimental setting of \cite{saragadam2023wire}, which sample points over a $512 \times 512 \times 512$ grid.
We establish MLP with two hidden layers and each hidden layer contains 256 neurons. We reparameterize the weight matrices between consecutive hidden layers and set $F=256,P=8$.
We use Adam optimizer to minimize the $\ell_2$ loss between sampled voxel values and INR approximations.
We train all the networks for 200 epoches with an initial learning rate of $5 \times 10^{-3}$.
The learning rate is reduced exponentially during the training phase to $5 \times 10^{-4}$ in the end of training.

\hspace{1pc}In Fig. \ref{Thai_state}, we visualize the shape representation results by different methods, the intersection over union (IOU) metrics are also provided for reference. 
Our Fourier reparameterization method is able to represent intricate geometric shape with less artifacts and more details.

\subsection{Learning neural radiance fields}
\label{Nerf}
\hspace{1pc}Learning neural radience fields for view synthesis is also a main application of INR \cite{nerf1,nerf2,nerf3}. 
The main process of the view synthesis task is to reconstruct 3D representation of an object from the given 2D images taken at various given angles. 
In this section, we evaluate our Fourier reparameterization method in this task.
The original NeRF and two recent SOTAs by neural networks, \textit{i.e.} the InstantNGP \cite{nerf2} and the DVGO \cite{nerf3}, are adopted. We follow the experimental settings of these works and train the models on the all objects of the Blender dataset \cite{nerf1} with $800 \times 800 $ resolutions. For the original NeRF, we reparameterize the weight matrices of last three hidden layers with $F=128, P=32$ and the “NeRF-pytorch” codebase \cite{lin2020nerfpytorch} is used. The detailed Fourier reparameterization settings for the InstantNGP and DVGO and complete results can be found in our supplementary file.

\begin{figure*}
    \centering
    \includegraphics[width=\linewidth]{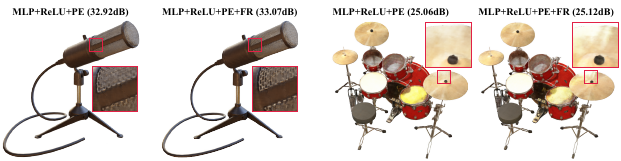}
    \caption{Visual examples of the view synthesis results (PSNR) by learning neural radiance field with the original NeRF \cite{nerf1}. }
    \label{fig:lego}
\end{figure*}

\hspace{1pc}In Fig. \ref{fig:lego}, we show some view synthesis results of the original NeRF. With our proposed Fourier reparameterization, the INR is able to capture more details of the complex texture, also representing the more accurate reflection of light, and therefore achieve better PSNR values.

\subsection{Ablation study}
\hspace{1pc}In this part, we conduct ablation studies to analyze our design choices.
All the ablation experiments are conducted on the first three images of the Kodak 24 dataset, and the network architectures are the same as our experimental settings in section \ref{2D_image}.

\subsubsection{Weight reparameterization with Fourier basis}

\label{ablation1}
\hspace{1pc}In Theorem 2, we show that appropriate bases selection could alleviate the low-frequency bias issue.
In this paper, we select fixed Fourier bases to reparameterize our MLP and show its superiority approximation accuracy on several vision applications.
In this section, we conduct experiments to show that the selection of basis plays a crucial role in our approach.
We generate random basis from a uniform distribution and denote the model as random reparameterization (RR).
Moreover, we also adopt a similar strategy as the existing reparameterization works \cite{ding2021repvgg} which updates all the random initialized parameters instead of fixed basis during the training phase, and denote the model as 
randomly initializaed reparameterization (RIR). 
The approximation results by different reparameterization schemes are shown in Table \ref{tab:basis}.
The results clearly show that bases play a pivotal role in reparameterized training, and our selected Fourier bases have advantages in capturing fine details in the target function.

\begin{table}[]
    \centering
    \caption{Ablation experiments on the reparameterization bases. Our Fourier bases achieve the best appriximation results (PSNR) on all the three network architectures. Detailed experimental results can be found in Section \ref{ablation1}.}
    \scriptsize
    \begin{tabular}{l|ccc|c}
    \toprule
       Method & Kodim 01 & Kodim 02 &Kodim 03  & Average\\
    \midrule
    MLP+ReLU+RR &19.64 &26.19 &25.66 & 23.83\\
    MLP+ReLU+RIR   &  20.22& 26.48 &26.67 & 24.46 \\
      MLP+ReLU+FR &20.34 &26.58 &27.21& 24.71\\
      \midrule
    MLP+ReLU+PE+RR &27.05 &32.16 &33.38 & 30.86\\
    MLP+ReLU+PE+RIR &26.91  &32.04 &32.85& 30.60\\
    MLP+ReLU+PE+FR &27.64 &33.92 &34.45 & 32.00\\
      \midrule
    MLP+Sin+RR &34.28 &37.84 & 30.48 & 34.20\\
    MLP+Sin+RIR & 25.19  & 28.61  & 27.12 & 26.97\\
    MLP+Sin+FR & 33.45 &38.68 &39.58 & 37.24\\
      \bottomrule
    \end{tabular}
    \label{tab:basis}
\end{table}

\subsubsection{Training speed}
\label{training speed}
\hspace{1pc}As shown in Fig. \ref{fig:spectrum and loss}, in terms of learning iterations, our reparameterization method is able to accelerate the convergence speed of network training.
While, since our reparameterization method 
will lead to more computations in each training step, we present the detailed training time by different methods in the subsection.
In Table \ref{tab:time_usage}, we show the average per epoch training time by different methods in the simple function approximation experiment.
The additional time introduced by our Fourier reparameterizaiton is not significant.
Although it takes an additional $17.3 \%$ time (from $2.89\times 10^{-3}$ seconds to $3.31\times 10^{-3}$ seconds) for training a MLP + ReLu architecture,
our reparameterization method could lead to  $86 \%$ improvement 
on accuracy.

\begin{table}[]
    \centering
        \caption{Average per epoch training time by different methods in the simple function approximation experiment. Detaild experimental results can be found in Section \ref{experimental settings} and Section \ref{training speed}.}
    \scriptsize
    \begin{tabular}{l|cccc}
    \midrule
       Method  &   MLP+ReLU &MLP+ReLU+FR &MLP+Sin&MLP+Sin+FR\\
       \midrule
Time (ms) &2.89&3.31&3.50&  3.59\\
\bottomrule
    \end{tabular}
    \label{tab:time_usage}
\end{table}

\subsubsection{Sampling interval analysis}
\label{abla:length}
\hspace{1pc}Another important hyper-parameter for our method is our 
sampling interval.
In this section, we conduct experiments to analyze the effect of different sampling intervals.
Denote the maximum period of the adopted Fourier function as  $T_{max}$. We vary the sampling interval from $0.1 T_{max}$ to 
$10 T_{max}$.
The approximation accuracy with different sampling intervals can be found in Table \ref{Interval}.
Our method could achieve good results with a wide range of sampling intervals, \textit{i.e.} $0.5 T_{max}$ to $4 T_{max}$.
While sampling points from a very small range fails to represent an entire period for many bases and leads to a performance drop.

\begin{table}[]
    \centering
        \caption{Ablation experiments on sampling intervals. Our method could achieve good results (PSNR) on a wide range of sampling intervals. More experimental details can be found in section \ref{abla:length}.}
    \scriptsize
    \vspace{-\baselineskip}
    \begin{tabular}{l|ccc|c}
    \toprule
      Length& Kodim 01 & Kodim 02 &Kodim 03  & Average\\
    \midrule
    $T_{max}\times 0.1$&  19.23 & 25.69 &25.10 &  23.34\\
     $T_{max}\times 0.25$&  19.94& 26.28 &25.99 & 24.07\\
      $T_{max}\times 0.5$ &20.02 &26.82 &26.30& 24.38\\
      $T_{max}\times 1$ &20.08 &26.58 &27.21& 24.62\\
      $T_{max}\times 2$ &20.13&26.96 &27.05& 24.71\\
      $T_{max} \times 4$&20.01 &26.60 &26.99& 24.53\\
      $T_{max} \times 10$&19.47  & 25.73 &24.72&  23.14\\
    \bottomrule
    \end{tabular}
    \label{Interval}
    \vspace{-2\baselineskip}
\end{table}
\section{Conclusions}
\hspace{1pc}In this paper, we
proposed a novel Fourier reparameterization method for advanced implicit neural representation (INR).
We theoretically analyzed the low-frequency bias issue of multi-layer perceptron (MLP) for INR and show that appropriate network reparameterization is able to alleviate the low-frequency bias in training MLP.
Based on our theoretical analysis, we proposed our Fourier reparemeterization method which learns coefficient matrix of fixed Fourier bases to compose network weights instead of directly learning them from training data.
Experiments were conducted on simple function task and real-world vision applications.
Our method improved the representation accuracy for a wide range of commonly used INR network architectures.
We hope our initial study could inspire future works in adjusting the learning bias of network by advanced network parameterization.

{
    \small
\bibliographystyle{ieeenat_fullname}
    \bibliography{main}
}

\clearpage
\setcounter{page}{1}

\maketitlesupplementary

\vspace{3mm}
\hspace{1pc}In this file, we provide detailed proof of Theorem 2 in the main text, derivations of our initialization scheme and more experimental results on various tasks.
We provide detailed proof of our Theorem 2 in section \ref{proof of theorem 2} and present the derivations of our  
initialization scheme in section \ref{initialization}.
A detailed ablation study on our hyper-parameters: frequency number and phase number, is presented in \ref{design choices}.
In section \ref{image fitting on more activation}, we provide 2D image approximation results of more activation functions and input adjustment techniques and show more visual examples. 
In section \ref{3D on more activation}, we present more experimental results on the shape representation task.
Lastly, in section \ref{nerf on more scenes}, we show more view synthesis results by our method and provide the full results of three NeRF frameworks \cite{nerf1,nerf2,nerf3}.

\appendix
\section{Detailed proof of Theorem 2}
\label{proof of theorem 2}
\hspace{1pc}Recall that we define $\mathbb{L}(k)$ as the loss function at frequency $k$:
\begin{equation}
    \mathbb{L}(k)=|\mathcal{F}[f_{\mathbf{\Theta}}](k)-\mathcal{F}[g](k)|^2.
\end{equation}
Then we have the following Theorem 2.
\makeatletter
\def\@begintheorem#1#2{\trivlist
\item[\hskip \labelsep{\bfseries #1\ }]\itshape}
\makeatother

\textbf{Theorem 2.}
Given a MLP with multiple hidden layers, reparameterize the weight matrix $\mathbf{W}\in \mathbb{R}^{d\times d}$ of one hidden layer with a trainable coefficient matrix $\mathbf{\Lambda}\in \mathbb{R}^{d\times M}$ and the fixed basis matrix $\mathbf{B} \in \mathbb{R}^{M\times d}$. For any frequencies $k_1$ and $k_2$ such that $k_1>k_2>0$, given any $\epsilon \geq 0$ and fixed $i$, for $ j=1,2,\dots,M$, there must exist a set of basis matrices such that 
\begin{equation}
    |\frac{\partial \mathbb{L}(k_1)}{\partial \lambda_{ij}}/ \frac{\partial \mathbb{L}(k_2)}{\partial \lambda_{ij}}| \geq \max \{|\frac{\partial \mathbb{L}(k_1)}{\partial w_{i1}}/\frac{\partial \mathbb{L}(k_2)}{\partial w_{i1}}|,\dots,|\frac{\partial \mathbb{L}(k_1)}{\partial w_{id}}/\frac{\partial \mathbb{L}(k_2)}{\partial w_{id}}|\}-\epsilon,
\end{equation}
    \label{theorem2}
    where $\mathbf{W}(i,j)=w_{ij}$ and $\mathbf{\Lambda}(i,j)=\lambda_{ij}$.
\begin{proof}
Before the detailed proof, simply denote: $\mathbb{L}_{\lambda_{ij}}(k_1)=\frac{\partial \mathbb{L}(k_1)}{\partial \lambda_{ij}},\mathbb{L}_{w_{ij}}(k_1)=\frac{\partial \mathbb{L}(k_1)}{\partial w_{ij}} $.

First, the weight reparameterization for $\mathbf{W}$ is expressed as follows:
\begin{equation}
    \mathbf{W}=\mathbf{\Lambda B},
\end{equation}
by the matrix multiplication, for any $w_{ij}\in \mathbf{W}$, the follow equation holds true:
\begin{equation}
    w_{ij}=\begin{bmatrix}
        \lambda_{i1},\lambda_{i2},\dots ,\lambda_{iM}
    \end{bmatrix}\begin{bmatrix}
        b_{1j}\\
        b_{2j}\\
        \vdots\\
        b_{Mj}
    \end{bmatrix},
\end{equation}
where $B(i,j)=b_{ij}$. 
Regarding $w_{i1},\dots,w_{id}$ as the latent variables related with $\lambda_{ij}$, for all $\lambda_{ij} \in \mathbf{\Lambda}$, using the chain rule, we have the following relationships:

\begin{equation}
    \mathbb{L}_{\lambda_{ij}}(k)=\sum_{t=1}^{d} b_{jt} \mathbb{L}_{w_{it}}(k).
    \label{equation_freq}
\end{equation}

Second, given two frequecies $k_1>k_2>0$, for the $i$-th row of $ \mathbf{\Lambda}$, we set that:

\begin{equation}
     \tau =arg \max_{j} \{|\mathbb{L}_{w_{ij}}(k_1)/\mathbb{L}_{w_{ij}}(k_2)| \}.
\end{equation}

Further, considering the elements of $\mathbf{B}$, for $j=1,\dots, M$, we make $|b_{jt}|< \alpha$ for $ t \neq \tau$ and $b_{j\tau}=1$. $\alpha$ is a positive upper bound. 
Then, according to equation \ref{equation_freq}, for the fixed $i$, for $j=1,\dots,M$, we have:

\begin{equation}
    \begin{split}
        \mathbb{L}_{\lambda_{ij}}(k_1)=\mathbb{L}_{w_{i\tau}}(k_1)+\sum_{t\neq \tau}b_{jt} \mathbb{L}_{w_{it}}(k_1)\\
         \mathbb{L}_{\lambda_{ij}}(k_2)=\mathbb{L}_{w_{i\tau}}(k_2)+\sum_{t\neq \tau}b_{jt} \mathbb{L}_{w_{it}}(k_2).
    \end{split}
\end{equation}

We denote $G_1$ and $G_2$ as $\sum_{t\neq\tau}|\mathbb{L}_{w_{it}}(k_1)|$ and $\sum_{t\neq\tau}|\mathbb{L}_{w_{it}}(k_2)|$, respectively.

\hspace{1pc}Without loss of generality, for any 
\begin{equation}
    0\leq\epsilon\leq \frac{|\mathbb{L}_{w_{i\tau}}(k_1)|}{|\mathbb{L}_{w_{i\tau}}(k_2)|},
\end{equation}
set 
\begin{equation}
\begin{split}
    \alpha \leq \min \{\frac{|\mathbb{L}_{w_{i \tau }}(k_2)|\epsilon}{G_1+G_2|\mathbb{L}_{w_{i \tau }}(k_1)/ \mathbb{L}_{w_{i \tau }}(k_2)|-G_2 \epsilon},
    |\frac{\mathbb{L}_{w_{i\tau }}(k_1)}{G_1}|\},
\end{split}
\label{alpha}
\end{equation}
then by inequalities involving absolute values, we have:
\begin{equation}
    \begin{split}
        |\mathbb{L}_{\lambda_{ij}}(k_1)/ \mathbb{L}_{\lambda_{ij}}(k_2)|
        =|\frac{\mathbb{L}_{w_{i\tau}}(k_1)+\sum_{t \neq \tau }b_{jt}\mathbb{L}_{w_{it}}(k_1)}{\mathbb{L}_{w_{i\tau}}(k_2)+\sum_{t \neq \tau }b_{jt}\mathbb{L}_{w_{it}}(k_2)}|
        \geq \frac{||\mathbb{L}_{w_{i\tau}}(k_1)|-|\sum_{t\neq \tau }b_{jt} \mathbb{L}_{w_{it}}(k_1)||}{|\mathbb{L}_{w_{i\tau}}(k_2)|+|\sum_{t \neq \tau }b_{jt}\mathbb{L}_{w_{it}}(k_2)|}.
    \end{split}
\end{equation}
From \ref{alpha}, we have that $\alpha \leq |\frac{\mathbb{L}_{w_{i\tau }}(k_1)}{G_1}|$. Then:
\begin{equation}
    |\sum_{t \neq \tau }b_{jt}\mathbb{L}_{w_{it}}(k_1)|\leq 
    \alpha \sum_{t\neq\tau}|\mathbb{L}_{w_{it}}(k_1)|=\alpha G_1 \leq 
    |\mathbb{L}_{w_{i\tau }}(k_1)|,
\end{equation}
which means that $|\mathbb{L}_{w_{i\tau}}(k_1)|-|\sum_{t\neq \tau }b_{jt}\mathbb{L}_{w_{it}}(k_1)| \geq 0$. Thus, the following inequalities holds true:
\begin{equation}
    \begin{split}
    \frac{|\mathbb{L}_{w_{i\tau}}(k_1)|-|\sum_{t\neq \tau }b_{jt} \mathbb{L}_{w_{it}}(k_1)|}{|\mathbb{L}_{w_{i\tau}}(k_2)|+|\sum_{t \neq \tau }b_{jt}\mathbb{L}_{w_{it}}(k_2)|}
        \geq \frac{|\mathbb{L}_{w_{i\tau}}(k_1)|-\sum_{t\neq \tau }|b_{jt} ||\mathbb{L}_{w_{it}}(k_1)|}{|\mathbb{L}_{w_{i\tau}}(k_2)|+\sum_{t \neq \tau }|b_{jt}||\mathbb{L}_{w_{it}}(k_2)|}
        \geq \frac{|\mathbb{L}_{w_{i\tau}}(k_1)|-\alpha G_{1}}{|\mathbb{L}_{w_{i\tau}}(k_2)|+ \alpha G_{2}} \geq 0
    \end{split}
\end{equation}
Substituting $\alpha \leq \frac{|\mathbb{L}_{w_{i \tau }}(k_2)|\epsilon}{G_1+G_2|\mathbb{L}_{w_{i \tau }}(k_1)/ \mathbb{L}_{w_{i \tau }}(k_2)|-G_2 \epsilon}$ into the above inequality, for the fixed $i$ and for $j=1 ,\dots, M$, we have that:
\begin{equation}
\begin{split}
    |\mathbb{L}_{\lambda_{ij}}(k_1)/ \mathbb{L}_{\lambda_{ij}}(k_2)|\geq& \frac{|\mathbb{L}_{w_{i\tau}}(k_1)|-\alpha G_{1}}{|\mathbb{L}_{w_{i\tau}}(k_2)|+ \alpha G_{2}}\\
\geq & \frac{|\mathbb{L}_{w_{i\tau}}(k_1)|(G_1+G_2|\mathbb{L}_{w_{i \tau }}(k_1)/ \mathbb{L}_{w_{i \tau }}(k_2)|-G_2 \epsilon)-G_1 |\mathbb{L}_{w_{i \tau }}(k_2)| \epsilon}{|\mathbb{L}_{w_{i\tau}}(k_2)|(G_1+G_2|\mathbb{L}_{w_{i \tau }}(k_1)/ \mathbb{L}_{w_{i \tau }}(k_2)|-G_2 \epsilon)+G_2|\mathbb{L}_{w_{i \tau }}(k_2)| \epsilon}\\
    =& \frac{G_1|\mathbb{L}_{w_{i\tau}}(k_1)|+\frac{G_2 |\mathbb{L}_{w_{i\tau}}(k_1)|^2}{|\mathbb{L}_{w_{i\tau}}(k_2)|}}{G_1|\mathbb{L}_{w_{i\tau}}(k_2)|+G_2|\mathbb{L}_{w_{i\tau}}(k_1)|}-\epsilon\\
    =& |\frac{\mathbb{L}_{w_{i\tau}}(k_1)}{\mathbb{L}_{w_{i\tau}}(k_2)}||\frac{G_1+\frac{G_2 |\mathbb{L}_{w_{i\tau}}(k_1)|}{|\mathbb{L}_{w_{i\tau}}(k_2)|}}{G_1+G_2\frac{|\mathbb{L}_{w_{i\tau}}(k_1)|}{|\mathbb{L}_{w_{i\tau}}(k_2)|}}|-\epsilon \\
    =& |\frac{\mathbb{L}_{w_{i\tau}}(k_1)}{\mathbb{L}_{w_{i\tau}}(k_2)}|-\epsilon\\
    =& \max \{ \frac{|\mathbb{L}_{w_{i1}}(k_1)|}{|\mathbb{L}_{w_{i1}}(k_2)|},\dots,\frac{|\mathbb{L}_{w_{id}}(k_1)|}{|\mathbb{L}_{w_{id}}(k_2)|}\}-\epsilon.
\end{split}
\end{equation}
\end{proof}

\section{Initialization scheme}
\label{initialization}
\hspace{1pc}Recall that we consider the initialization of the coefficient matrix $\mathbf{\Lambda^{(n)}}\in \mathbb{R}^{d_{n} \times M}$. Inspired by Kaiming initialization \cite{multi-branch_1}, the initialization scheme of the $i$-th row in $\mathbf{\Lambda^{(n)}}$ should satisfy:
\begin{equation}
Var(\boldsymbol{w^{(n)}_{i}}\boldsymbol{x})=Var(\mathbf{\boldsymbol{\lambda^{(n)}_{i}}}\mathbf{B^{(n)}}\boldsymbol{x}),
\label{condition}
\end{equation}
where $\boldsymbol{w^{(n)}_{i}} \in \mathbb{R}^{1 \times d_{n-1}}$ and $\boldsymbol{\boldsymbol{\lambda}^{(n)}_i}\in \mathbb{R}^{1 \times M}$ are the $i$-th row of the weights matrix $\mathbf{W^{(n)}}\in \mathbb{R}^{d_n \times d_{n-1}}$ and the coefficient matrix $\mathbf{\Lambda^{(n)}}$, respectively; $Var(\cdot)$ denotes the variance; $\boldsymbol{x}\in \mathbb{R}^{d_{n-1}\times 1}$ is the input of this layer; $\mathbf{B^{(n)}}\in \mathbb{R}^{M \times d_{n-1}}$ is the fixed basis matrix. We assume that the elements of  $\mathbf{W^{(n)}}, \mathbf{\Lambda^{(n)}}$ and the bias vector $\boldsymbol{b^{(n)}}$ are statistically independent of each other. Therefore we can omit the bias vector $\boldsymbol{b^{(n)}}$ on the variance. We assume that the outputs of different neurons in each layer of the neural network are independent. Then the left-hand side of equation \ref{condition} expands as follows:
\begin{equation}
        Var(\boldsymbol{w^{(n)}_{i}}\boldsymbol{x})= Var(\sum_{j=1}^{d_{n-1}} w^{(n)}_{ij}x_j)= \sum_{j=1}^{d_{n-1}}Var(w^{(n)}_{ij}x_j),
    \label{expands1}
\end{equation}
where $x_j$ is the $j$-th element of $\boldsymbol{x}$. We let $x_j$ have the same distribution for $j=1,\dots,d_{n-1}$ accroding to Kaiming initialization \cite{multi-branch_1}. Then, the equation \ref{expands1} can be replaced by $\sum_{j=1}^{d_{n-1}}Var(w^{(n)}_{ij}x_1)$:
\begin{equation}
    Var(\boldsymbol{w^{(n)}_{i}}\boldsymbol{x})=Var(x_1\sum_{j=1}^{d_{n-1}}w^{(n)}_{ij}).
\end{equation}
Similarly, for the right-hand side, we also have that:
\begin{equation}
Var(\mathbf{\boldsymbol{\lambda^{(n)}_{i}}}\mathbf{B^{(n)}}\boldsymbol{x})=Var(x_1\sum_{t=1}^{d_{n-1}}\sum_{j=1}^{M}\lambda^{(n)}_{ij}b^{(n)}_{jt}).
\end{equation}
As the following equation holds true \cite{Variance}:
\begin{equation}
Var(XY)=Var(X)Var(Y)+(E(X))^2Var(Y)+(E(Y))^2Var(x),
\end{equation}
where $X,Y$ are the independent random variables and $E( \cdot )$ denotes the mathematical expectation. 

By this, we further expand equation \ref{condition} as follows:
\begin{equation}
    \begin{split}
        &Var(\boldsymbol{w^{(n)}_{i}}\boldsymbol{x})
        =Var(x_1)Var(\sum_{j=1}^{d_{n-1}}w^{(n)}_{ij})+(E(x_1))^2Var(\sum_{j=1}^{d_{n-1}}w^{(n)}_{ij})
        +(E(\sum_{j=1}^{d_{n-1}}w^{(n)}_{ij}))^2 Var(x_1),\\
    &Var(\boldsymbol{\lambda^{(n)}_{i}}\mathbf{B^{(n)}}\boldsymbol{x})=Var(x_1)Var(\sum_{t=1}^{d_{n-1}}\sum_{j=1}^{M}\lambda^{(n)}_{ij}b^{(n)}_{jt})+(E(x_1))^2Var(\sum_{t=1}^{d_{n-1}}\sum_{j=1}^{M}\lambda^{(n)}_{ij}b^{(n)}_{jt})+(E(\sum_{t=1}^{d_{n-1}}\sum_{j=1}^{M}\lambda^{(n)}_{ij}b^{(n)}_{jt}))^2 Var(x_1).
    \end{split}
\end{equation}
Note that a sufficient condition to make equation \ref{condition} hold true is that $Var(\sum_{t=1}^{d_{n-1}}\sum_{j=1}^{M}\lambda_{ij}b^{(n)}_{jt})=Var(\sum_{j=1}^{d_{n-1}}w^{(n)}_{ij})$ and $E(\lambda^{(n)}_{ij})=0$. Hence, we let $\lambda^{(n)}_{ij} \sim U(-a,a),a>0$ to satisfy that $E(\lambda^{(n)}_{ij})=0$, which $a$ is a parameter to be determined. And $Var(\lambda_{ij}^{(n)})=\frac{a^2}{3}$. Further, from the above variance constraint, we have that:
\begin{equation}
Var(\sum_{t=1}^{d_{n-1}}\sum_{j=1}^{M}\lambda^{(n)}_{ij}b^{(n)}_{jt})=\sum_{t=1}^{d_{n-1}}\sum_{j=1}^M {{b^{(n)}_{jt}}}^2 Var(\lambda^{(n)}_{ij})=
\sum_{j=1}^M \sum_{t=1}^{d_{n-1}} {{b^{(n)}_{jt}}}^2 Var(\lambda^{(n)}_{ij})
=\sum_{j=1}^M (\sum_{t=1}^{d_{n-1}} {{b^{(n)}_{jt}}}^2) Var(\lambda^{(n)}_{ij}).
\end{equation}
We simply make $(\sum_{t=1}^{d_{n-1}} {{b^{(n)}_{jt}}}^2) Var(\lambda^{(n)}_{ij})=\frac{Var(\sum_{j=1}^{d_{n-1}}w^{(n)}_{ij})}{M}$. We can have that :
\begin{equation}
    Var(\lambda^{(n)}_{ij})=\frac{Var(\sum_{j=1}^{d_{n-1}}w^{(n)}_{ij})}{M\sum_{t=1}^{d_{n-1}} 
    {b^{(n)}_{jt}}^2}=\frac{a^2}{3}
\end{equation}
When $w^{(n)}_{ij} \sim U(-\sqrt{\frac{6}{d_{n-1}}},\sqrt{\frac{6}{d_{n-1}}} )$, we have that $a= \sqrt{\frac{6}{M \sum_{t=1}^{d_{n-1}}{{{b^{(n)}_{jt}}}^2}}}$.
Thus, the following initialization scheme for ReLU activation function can be obtained as follows:
\begin{equation}
    \lambda^{(n)}_{ij}\sim U(-\sqrt{\frac{6}{M \sum_{t=1}^{d_{n-1}}{{b^{(n)}_{jt}}}^2}},\sqrt{\frac{6}{M \sum_{t=1}^{d_{n-1}}{{b^{(n)}_{jt}}}^2}} ). 
\end{equation}

The initialization scheme for other activation function can be obtained by the similar deduction. As for the parameters without reparameterization, the initialization scheme is remained unchanged.

\section{Design choices analysis}
\label{design choices}
\begin{figure}
    \centering
    \includegraphics[width=0.3\linewidth]{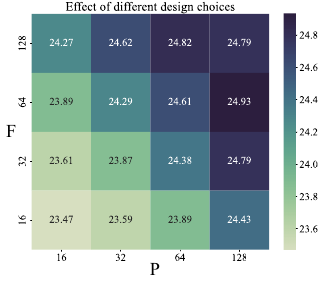}
    \caption{Visualization of the effect from differnet design choices (x-axis for phase number, y-axis for frequency number and colormap for PSNR value). More experimental details can be found in the Section \ref{design choices}.}
    \label{fig:different_choice}
\end{figure}

\hspace{1pc}As discussed in the main text, our Fourier reparameterization method has two hyper-parameters: frequency number $F$ and phase number $P$. 
In this section, we provide experimental results to show the effects of different combinations of  $F$ and $P$. 
We vary the frequency number $F$  and phase number $P$ from 16 to 128, the approximation accuracy by different design choices can be found in Fig. \ref{fig:different_choice}. The approximation accuracy is the average PSNR on the first 3 images of Kodak 24 dataset.

Generally, reparameterizing the weight matrix with more bases will lead to better approximation accuracy. Also, we need to have a balanced number of frequencies and phases to achieve good results. 

\section{2D Color image approximation for more activation functions and input adjustment techniques }
\label{image fitting on more activation}
\hspace{1pc}In the main text, we evaluated our Fourier reparameterization method on MLP+ReLU, MLP+ReLU+PE and MLP+Sin. 
In this section, we apply our Fourier reparameterization method to more activation functions, \textit{i.e.} the Tanh, the Gauss \cite{guass} and the Garbor wavelet \cite{saragadam2023wire} activation functions. The advanced input adjustment technique, \textit{i.e.} DINER \cite{diner}, is also applied with our method.
We follow the previous MLP structures and Fourier basis settings. The same training strategy is adopted for these experiments. For DINER coupled with Sin activation function, we early stops at 3000 iterations as its fast convergence.
\begin{table*}
\centering
\hspace*{-1cm}
\vspace{-5mm}
\footnotesize
\caption{Peak signal to noise ratio (PSNR) of 2D color image approximation results by different methods. MLP+Gauss denotes the MLP with Gauss activation function \cite{guass}. MLP+CGW denotes the MLP equipped with complex Gabor wavelet activation function \cite{saragadam2023wire}. MLP+ReLU+DINER denotes the MLP+ReLU coupled with the adjusted input features by a hash-table \cite{diner}. Detailed experiment settings can be found in Section \ref{2D_image} and Appendix \ref{image fitting on more activation}.}
\begin{tabular}{l|cccccccc|c}
\toprule
Method & Kodim 01 & Kodim 02 & Kodim 03 & Kodim 04 & Kodim 05 & Kodim 06 & Kodim 07 & Kodim 08 & Average \\
\midrule
MLP + ReLU   & 19.37 & 26.12 & 25.11 & 24.57 & 17.31 & 21.69 & 20.79 & 15.68 & 21.33 \\
MLP + ReLU + FR & 20.34 & 26.58 & 27.21 & 25.72 & 18.33 & 22.25 & 22.47 & 16.64 & 22.44 \\
\midrule
MLP + ReLU + PE & 24.47 & 31.41 & 31.53 & 30.16 & 22.87 & 26.54 & 29.33 & 21.14 & 27.18 \\
MLP + ReLU + PE + FR & 27.64 & 33.92 & 34.45 & 33.23 & 26.78 & 29.83 & 34.13 & 24.70 & 30.59 \\
\midrule
MLP + Sin & 31.59 & 36.55 & 39.59 & 36.66 & 33.05 & 34.10 & 39.96 & 31.00 & 35.31 \\
MLP + Sin + FR & 33.45 & 38.68 & 39.58 & 37.96 & 34.64 & 34.45 & 39.76 & 32.16 & 36.34 \\
\midrule
MLP + Tanh   & 17.30 & 22.18 & 21.05 & 19.94 & 15.47 & 19.96 & 18.47 & 15.52 & 18.74 \\
MLP + Tanh + FR & 19.15 & 24.95 & 25.36 & 23.89 & 17.32 & 21.46 & 21.31 & 16.09 & 21.19 \\
\midrule
MLP + Gauss & 24.83 & 30.29 & 31.32 & 30.40 & 24.79 & 25.78 & 29.40 & 22.42 & 27.40 \\
MLP + Gauss + FR & 24.86 & 30.19 & 31.40 & 31.05 & 24.98 & 25.53 & 27.75 & 24.98 & 27.59 \\
\midrule
MLP + CGW & 26.53 & 32.18 & 32.60 & 31.97 & 25.96 & 28.29 & 32.19 & 23.70 & 29.18 \\
MLP + CGW + FR & 28.54 & 33.56 & 35.09 & 33.60 & 28.12 & 30.17 & 34.47 & 25.68 & 31.15 \\
\midrule
MLP + ReLU + DINER & 45.65 & 50.28 & 37.57 & 44.01 & 39.69 & 42.54 & 44.50 & 41.15 & 43.17 \\
MLP + ReLU + DINER + FR & 45.81 & 50.53 & 38.06 & 43.79 & 40.42 & 43.13 & 44.45 & 41.35 & 43.44\\
\midrule
MLP + Sin + DINER &45.00 & 50.36 & 37.84 & 41.83 & 40.76 & 44.50 & 44.47 & 41.62 & 43.30 \\
MLP + Sin + DINER + FR & 47.45 & 50.67 & 44.74 & 46.65 & 40.89 & 43.85 & 42.92 & 43.56 & 45.09 \\

\bottomrule
\end{tabular}
\label{Table:Kodak24_appendix}
\end{table*}
In Table \ref{Table:Kodak24_appendix}, we report the PSNR achieved by these three INRs, DINER and previous models for approximating the first 8 images in the Kodak 24 dataset. 
The same as our results in the main text, our 
 Fourier reparameterization consistently improves the approximation accuracy for all the evaluated activation functions and input adjustments techniques.
 Some visual examples of the learned approximations by different models can be found in the following Fig. \ref{fig:Koda_01}, \ref{fig:Koda_04}, \ref{fig:Koda_08}.

\section{Representing shapes for more activation functions and scenes}
\label{3D on more activation}
\begin{figure*}
    \centering
    \includegraphics[width=\linewidth]{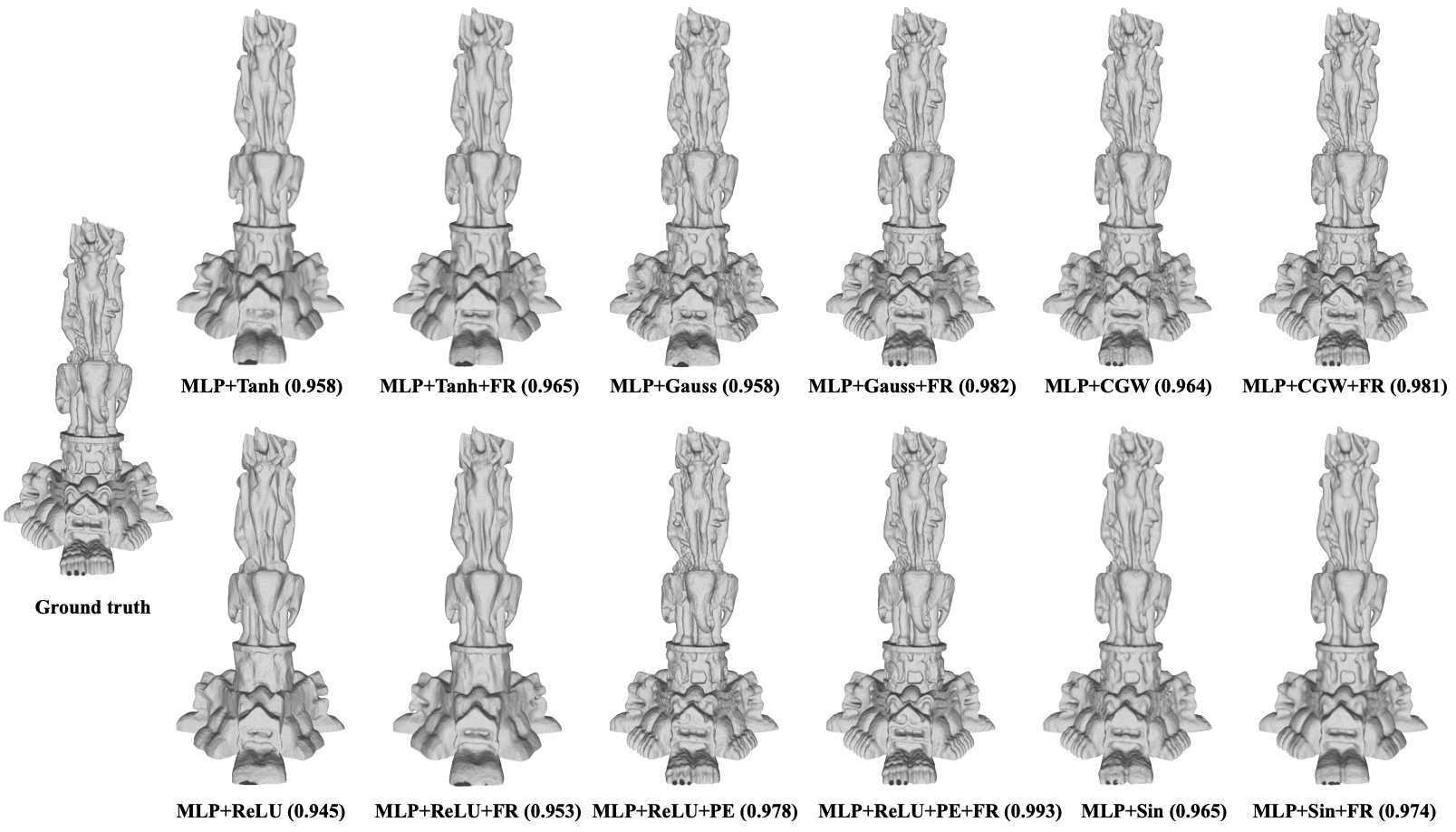}
    \caption{Visualization examples of the shape representation results (IOU) by different methods on Thai statue. Detailed experiment settings can be found in Section \ref{3D on more activation}.} 
    \label{fig:Thai_shape}
\end{figure*}

\begin{figure*}
    \centering
    \includegraphics[width=\linewidth]{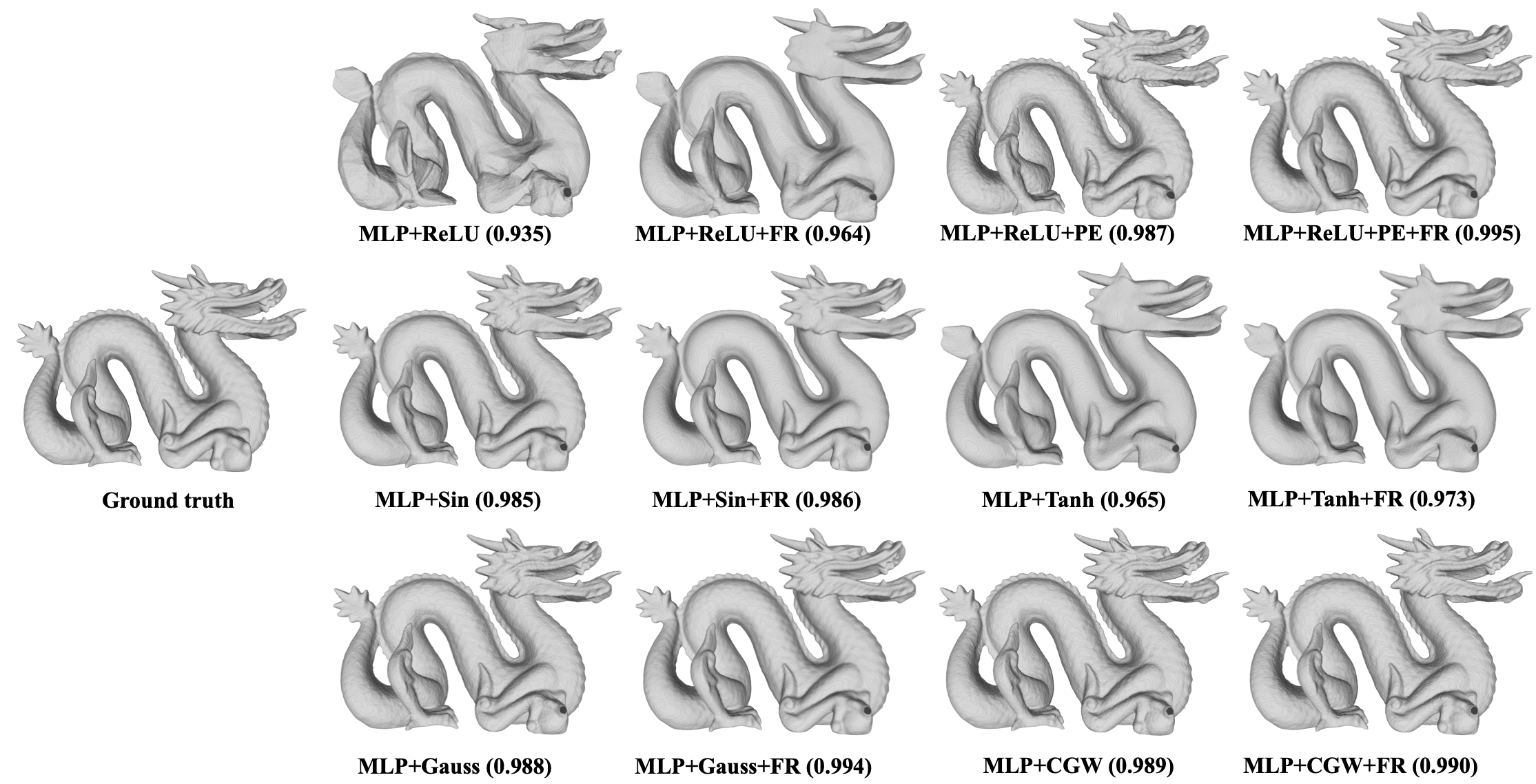}
    \caption{Visualization examples of the shape representation results (IOU) by different methods on Dragon Statue. Detailed experiment settings can be found in Section \ref{3D on more activation}.}
    \label{fig:Long_shape}
\end{figure*}

\hspace{1pc}Following the previous experimental settings and model structures, we further evaluate our Fourier reparameterization method with Tanh, Gauss \cite{guass} and complex Gabor wavelet \cite{saragadam2023wire} activation functions on the Thai statue and add the Dragon statue. Our Fourier reparameterization method helps models to capture more accurate complex shapes of the statue.
In Fig. \ref{fig:Thai_shape}, \ref{fig:Long_shape}, we visualize the shape representation results by these methods.

\section{Learning neural radiance fields for more scenes}
\label{nerf on more scenes}
\hspace{1pc}In the task of learning neural radiance fields, We evaluate our Fourier reparameterization method on the original NeRF \cite{nerf1} and two recent SOTAs with neural networks, \textit{i.e.} the DVGO \cite{nerf2} and the InstantNGP \cite{nerf3}. For the original NeRF, we set $F=128, P=32$. As for the small two-hidden-layer MLP of the DVGO and the InstantNGP, where the width of the hidden layers is 128 and 64, we reparameterize the weight matrix between consecutive hidden layers and empirically set $F=64, P=64$ and $F=32, P=64$ to ensure over-complete bases. The same experimental settings and training strategies as the original works are adopted.

\begin{figure*}
    \centering
    \includegraphics[width=0.9\linewidth]{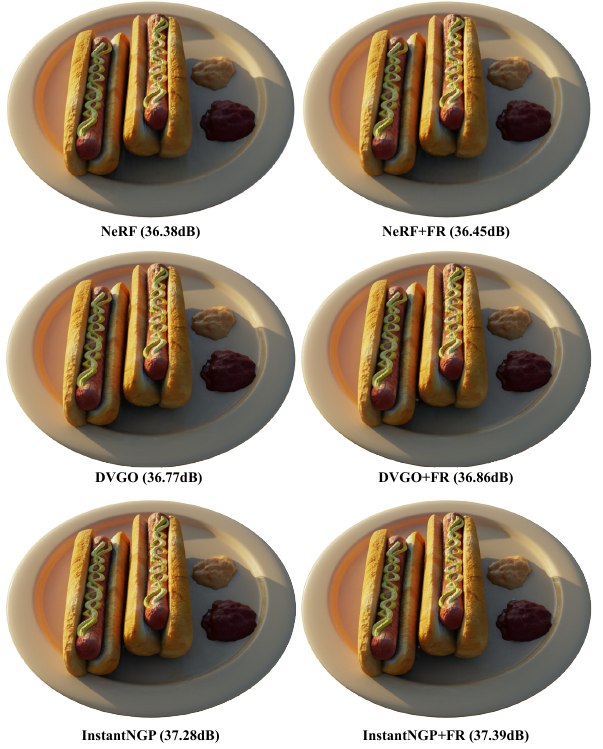}
    \caption{Visualization examples of the view synthesis results (PSNR) of “Hotdog” by learning neural radiance fields. Detailed experiment settings can be found in Section \ref{nerf on more scenes}.}
    \label{fig:nerf_other_two1}
\end{figure*}

\begin{figure*}
    \centering
    \includegraphics[width=0.8\linewidth]{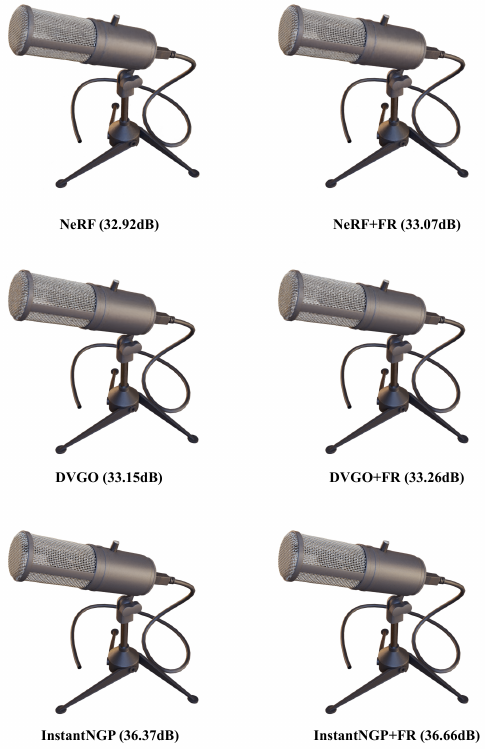}
    \caption{Visualization examples of the view synthesis results (PSNR) of “Mic” by learning neural radiance fields. Detailed experiment settings can be found in Section \ref{nerf on more scenes}.}
    \label{fig:nerf_other_two2}
\end{figure*}

\begin{figure*}
    \centering
    \includegraphics[width=0.9\linewidth]{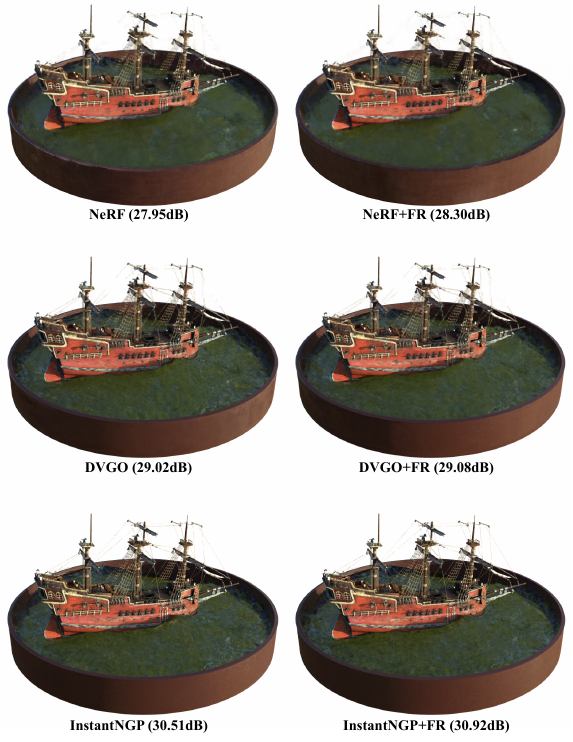}
    \caption{Visualization examples of the view synthesis results (PSNR) of “Ship” by learning neural radiance fields. Detailed experiment settings can be found in Section \ref{nerf on more scenes}.}
    \label{fig:nerf_other_two3}
\end{figure*}

\hspace{1pc}In Table \ref{Table:nerf}, we list the full results of three frameworks on the Blender dataset \cite{nerf1}. Our Fourier reparameterization method leads to more accurate view synthesis results. In Fig. \ref{fig:nerf_other_two1}, \ref{fig:nerf_other_two2}, \ref{fig:nerf_other_two3}, the detail of more reconstruction results is visualized. 

\begin{table*}
\centering
\footnotesize
\caption{Peak signal to noise ratio (PSNR) of view synthesis results by different methods on the Blender dataset \cite{nerf1}. NeRF+FR, DVGO+FR and InstantNGP+FR denote the frameworks of the original NeRF \cite{nerf1}, DVGO \cite{nerf2} and InstantNGP \cite{nerf3} trained with Fourier reparameterization. NeRF is reproduced on the “NeRF-pytorch” codebase \cite{lin2020nerfpytorch}. Detailed experiment settings can be found in Section \ref{Nerf} and Appendix \ref{nerf on more scenes}.} 
\begin{tabular}{l|cccccccc|c}
\toprule
Framework & Chair & Drums & Ficus & Hotdog & Lego & Materials  & Mic & Ship & Average \\
\midrule
NeRF \cite{nerf1}& 32.72 & 25.06 & 26.83 &36.38 & 32.55 & 29.55 & 32.92 & 27.95 & 30.50 \\
NeRF + FR (ours) & 32.73 & 25.12 & 30.15 & 36.45 & 32.59 & 29.56 & 33.07 & 28.30 &  31.00\\
\midrule
DVGO \cite{nerf3}& 34.07 & 25.39 & 32.66  & 36.77 & 34.65 & 29.59 & 33.15 & 29.02 & 31.91\\
DVGO + FR (ours) & 34.16 & 25.45 & 32.89 & 36.86 &34.78 & 29.73 & 33.26 & 29.08 & 32.03 \\
\midrule
InstantNGP \cite{nerf2} & 35.55 & 25.85 & 34.19 & 37.28 & 36.04 & 29.61 & 36.37 & 30.51 & 33.18\\
InstantNGP + FR (ours) & 35.64 & 25.88 & 34.23 & 37.39 & 36.10 & 29.63 & 36.66 & 30.92 & 33.31 \\
\bottomrule
\end{tabular}

\label{Table:nerf}
\end{table*}

\newpage
\begin{figure*}
    \centering
    \includegraphics[width=1\linewidth]{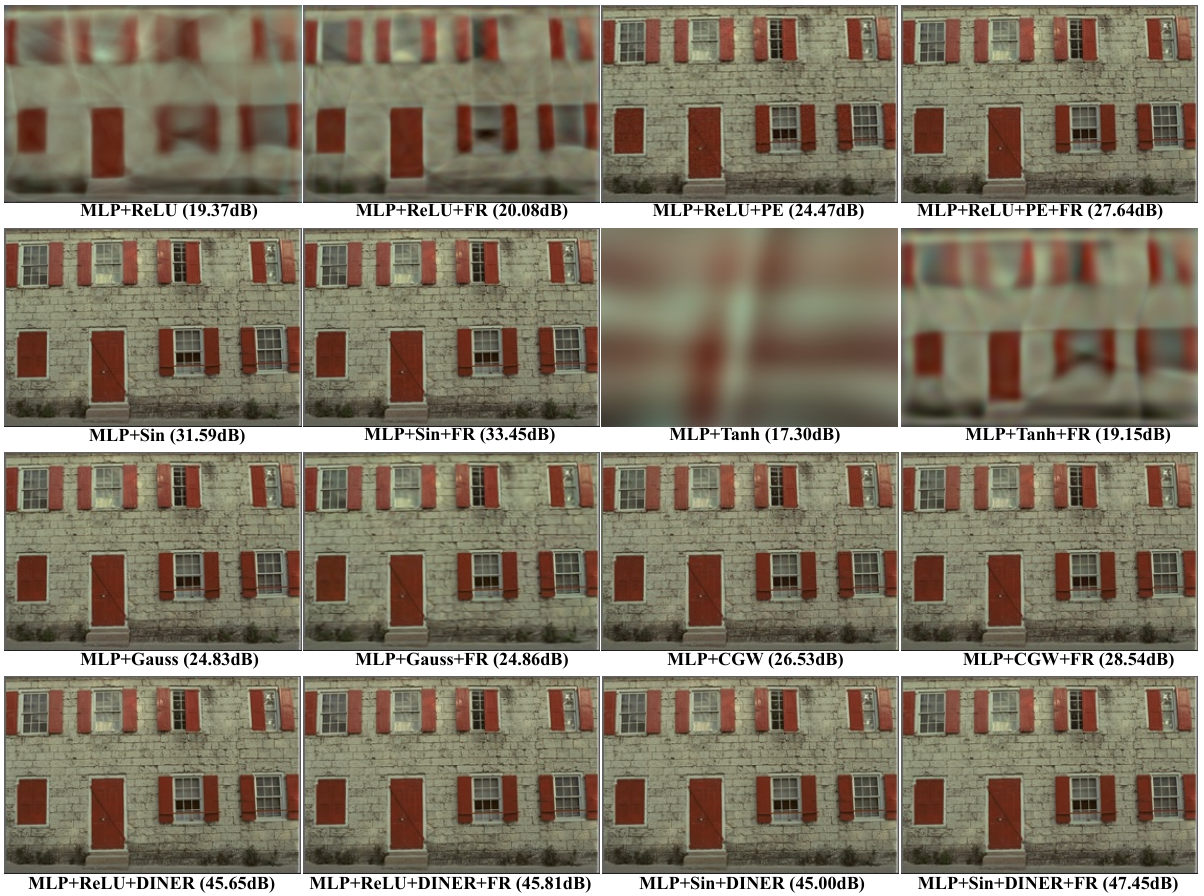}
    \caption{Peak signal to noise ratio (PSNR) of 2D color image approximation results on Kodim 01. MLP+Gauss denotes the MLP with Gauss activation function \cite{guass}. MLP+CGW denotes the MLP equipped with complex Gabor wavelet activation function \cite{saragadam2023wire}. MLP+ReLU+DINER denotes the MLP+ReLU coupled with the adjusted input features by a hash-table \cite{diner}. Detailed experiment settings can be found in Section \ref{2D_image} and Appendix \ref{image fitting on more activation}.}
    \label{fig:Koda_01}
\end{figure*}
\begin{figure*}
    \centering
    \includegraphics[width=1\linewidth]{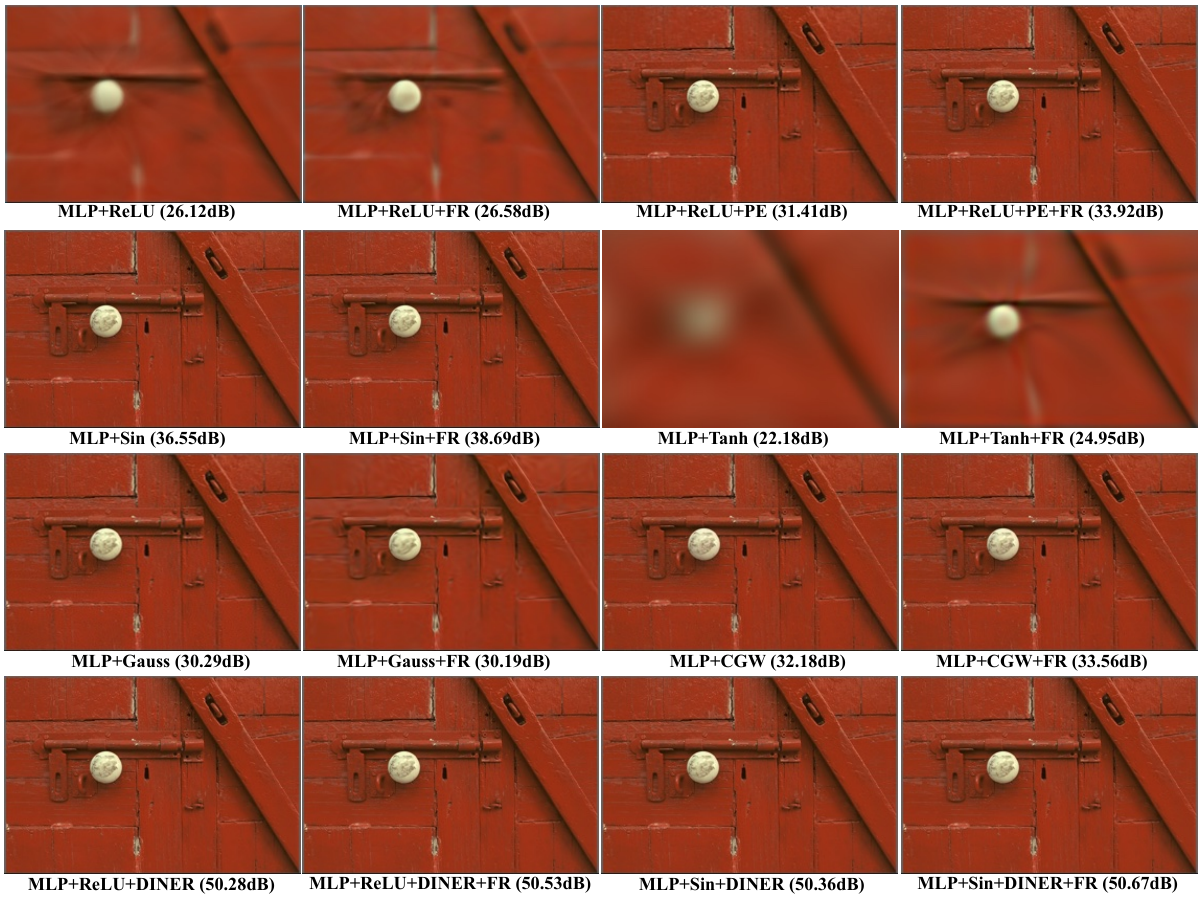}
    \caption{Peak signal to noise ratio (PSNR) of 2D color image approximation results on Kodim 02. MLP+Gauss denotes the MLP with Gauss activation function \cite{guass}. MLP+CGW denotes the MLP equipped with complex Gabor wavelet activation function \cite{saragadam2023wire}. MLP+ReLU+DINER denotes the MLP+ReLU coupled with the adjusted input features by a hash-table \cite{diner}. Detailed experiment settings can be found in Section \ref{2D_image} and Appendix \ref{image fitting on more activation}.}
    \label{fig:Koda_04}
\end{figure*}
\begin{figure*}
    \centering
    \includegraphics[width=1\linewidth]{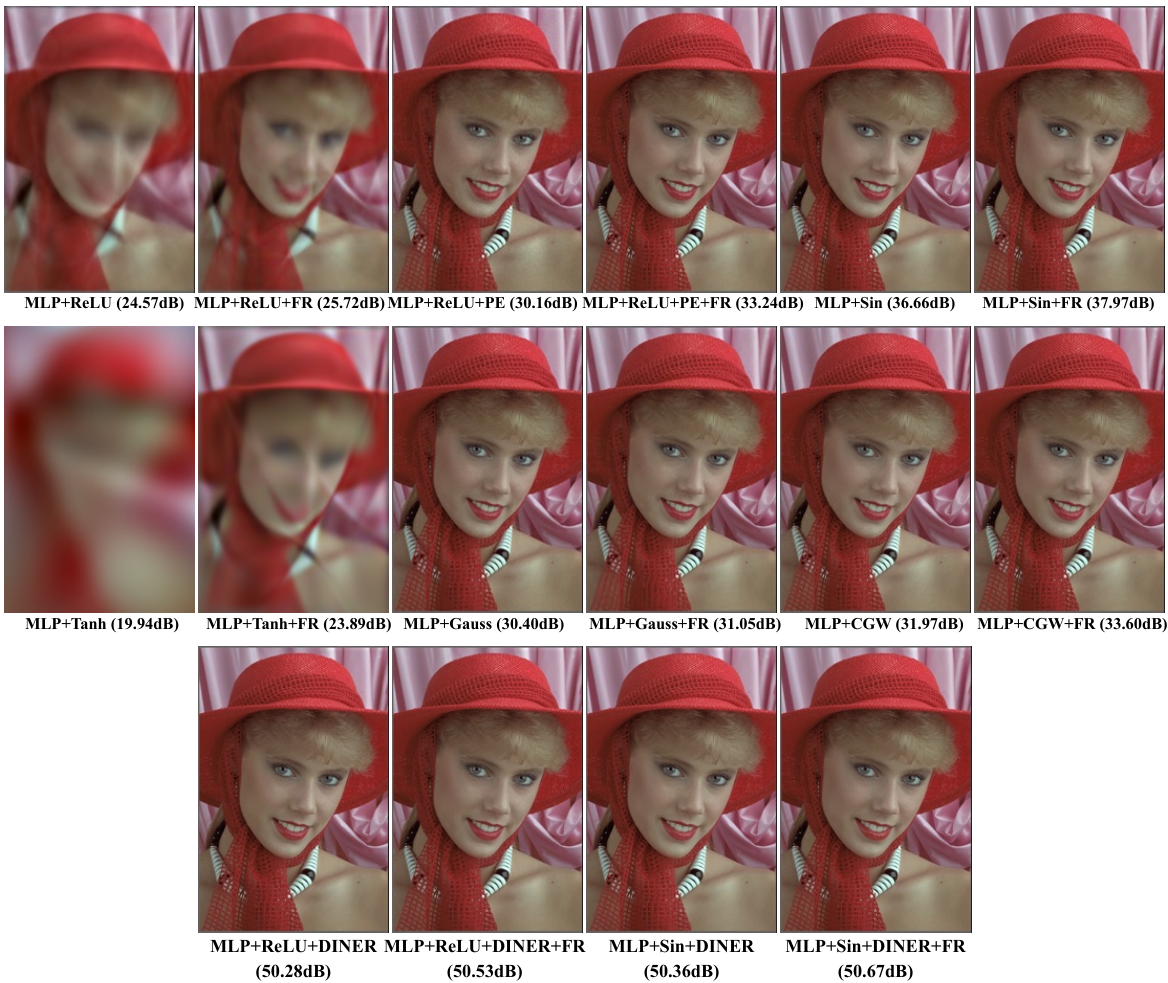}
    \caption{Peak signal to noise ratio (PSNR) of 2D color image approximation results on Kodim 04. MLP+Gauss denotes the MLP with Gauss activation function \cite{guass}. MLP+CGW denotes the MLP equipped with complex Gabor wavelet activation function \cite{saragadam2023wire}. MLP+ReLU+DINER denotes the MLP+ReLU coupled with the adjusted input features by a hash-table \cite{diner}. Detailed experiment settings can be found in Section \ref{2D_image} and Appendix \ref{image fitting on more activation}.}
    \label{fig:Koda_04}
\end{figure*}
\begin{figure*}
    \centering
    \includegraphics[width=1\linewidth]{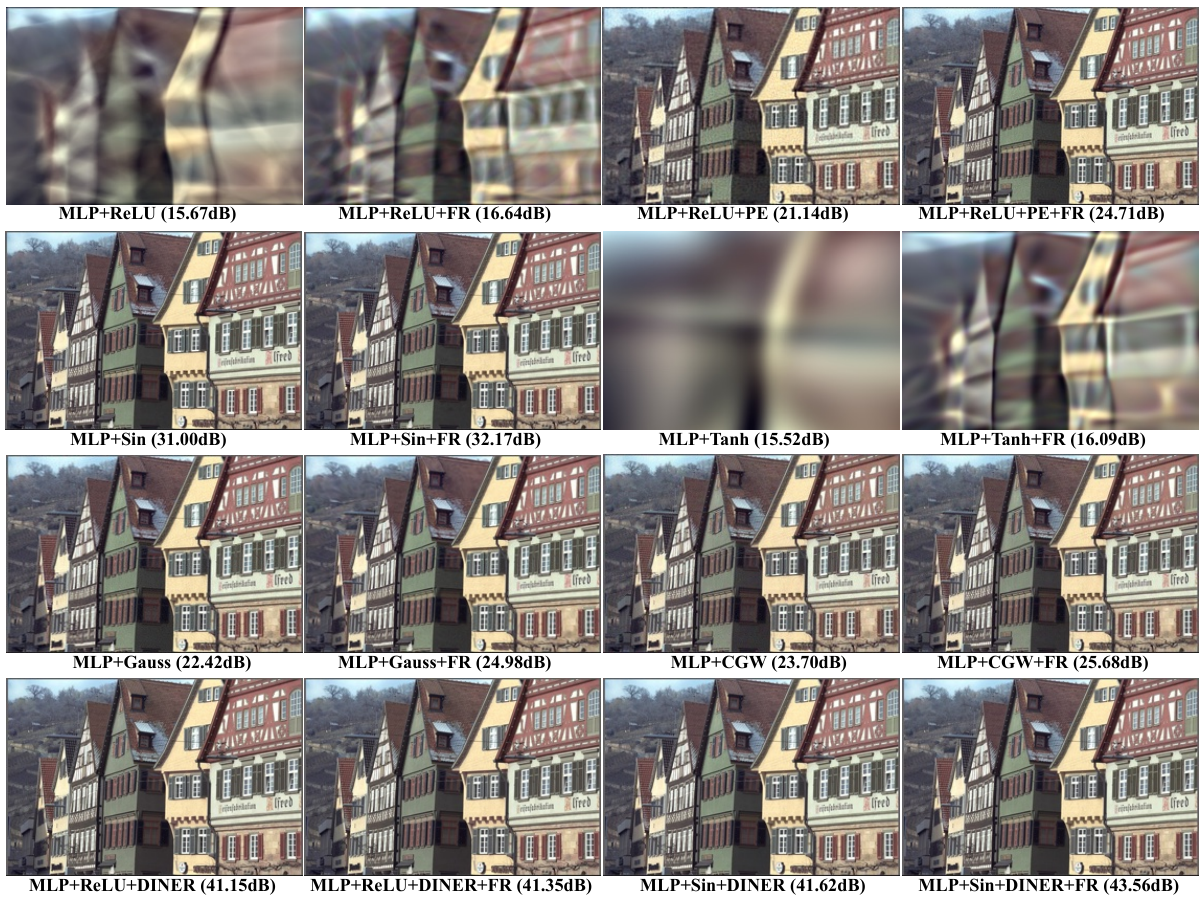}
    \caption{Peak signal to noise ratio (PSNR) of 2D color image approximation results on Kodim 08. MLP+Gauss denotes the MLP with Gauss activation function \cite{guass}. MLP+CGW denotes the MLP equipped with complex Gabor wavelet activation function \cite{saragadam2023wire}. MLP+ReLU+DINER denotes the MLP+ReLU coupled with the adjusted input features by a hash-table \cite{diner}. Detailed experiment settings can be found in Section \ref{2D_image} and Appendix \ref{image fitting on more activation}.}
    \label{fig:Koda_08}
\end{figure*}


\end{document}